\documentclass[preprint,12pt]{elsarticle}

\usepackage{graphicx}
\usepackage{siunitx}
\usepackage{subcaption}
\usepackage{makecell}
\usepackage{comment}
\usepackage{caption}
\usepackage{subcaption}
\usepackage{xurl}
\usepackage{hyperref}
\usepackage{breakurl}
\usepackage{url}
\usepackage{comment}
\usepackage{float}
\usepackage{hyperref}
\usepackage{array}
\usepackage{booktabs} 
\usepackage{caption}
\usepackage{subcaption}
\journal{Journal of Systems Architecture}

\begin{document}

\begin{frontmatter}




\title{Generative AI-based Pipeline Architecture for Increasing Training Efficiency in Intelligent Weed Control Systems} 


\author{Sourav Modak, Anthony Stein} 

\affiliation{organization={Department of Artificial Intelligence in Agricultural Engineering \& Computational
Science Hub, University of Hohenheim},
            addressline={Garbenstraße 9}, 
            city={Stuttgart},
            postcode={70599}, 
            country={Germany}}

\begin{abstract}

In automated crop protection tasks such as weed control, disease diagnosis, and pest monitoring, deep learning has demonstrated significant potential. However, these advanced models rely heavily on high-quality, diverse datasets, which are often scarce and costly to obtain in agricultural settings. Traditional data augmentation techniques, while useful for increasing the volume of the dataset, often fail to capture the real-world variability and conditions needed for robust model training. In this paper, we present a new approach for generating synthetic images for improved training of deep learning-based object detection models in the context of intelligent weed control. The presented approach is designed to improve the data efficiency of the model training process. The architecture of our GenAI-based image generation pipeline integrates the Segment Anything Model (SAM) for zero-shot domain adaptation with a text-to-image Stable Diffusion Model, enabling the creation of synthetic images that can accurately reflect the idiosyncratic properties and appearances of a variety of real-world conditions. We further assess the application of these synthetic datasets on edge devices by evaluating state-of-the-art lightweight YOLO models, measuring data efficiency by comparing mAP50 and mAP50\text{-}95 scores among different proportions of real and synthetic training data. Incorporating these synthetic datasets into the training process has been found to result in notable improvements in terms of data efficiency. For instance, most YOLO models that are trained on a dataset consisting of 10\% synthetic images and 90\% real-world images typically demonstrate superior scores on mAP50 and mAP50\text{-}95 metrics compared to those trained solely on real-world images. This shows that our approach not only reduces the reliance on large real-world datasets but at the same time also enhances the models' predictive performance. The integration of this approach opens opportunities for achieving continual self-improvement of perception modules in intelligent technical systems.

\end{abstract}




\end{frontmatter}
\section{Introduction}
\label{sec:Introduction}

In the transformation towards more sustainable agriculture, the adoption of smart technology for crop protection is crucial to minimize the application of pesticides. Here, deep learning (DL)-based algorithms embedded in intelligent agricultural technology systems emerge as a promising frontier in the area of automated crop protection~\cite{vithlani2023machine}. DL facilitates these tasks through a range of computer vision techniques, including image classification, object detection, and segmentation. A vast amount of data, however, is a prerequisite for the satisfactory performance of DL models. 

In agricultural scenarios, high-quality labeled open-source datasets reflecting the heterogeneity present, for instance, in the various field conditions are yet scarce. Moreover, collecting and annotating images for new datasets is a highly labor-intensive and expensive process~\cite{moreno2023analysis}.To alleviate this data scarcity issue, data augmentation is a popular tool in DL to increase both the volume of and variation from the available datasets~\cite{mumuni2022data}. 
Different approaches to data augmentation exist. 
Classical operations such as flipping, rotating, and blurring of images are straightforward to implement and use but lack the important property of introducing real variability into the training data. 
This is, however, important to yield robustly working models generalizing well to unseen scenes. 
Recent approaches such as those proposed in~\cite{incollectionIqbal} deal with generating artificial training data. 
We provide a brief overview in Section~\ref{sec:Background}.

This article expands upon our previous work presented at the Architecture of Computing Systems (ARCS) conference~\cite{stab_diff}, in which we developed a novel pipeline architecture for synthetic image generation adapted to weed detection, with applicability to other object detection tasks. Our approach combined the zero-shot transfer technique with the \textit{Segment Anything Model} (SAM)~\cite{kirillov2023segment} and the \textit{Stable Diffusion Model}~\cite{rombach2022high}. This integration aimed to generate synthetic image data that not only mimics the inherent style of real images but also enhances the natural variation within image datasets used to train deep neural network models. Validation was carried out using a sugar beet weed detection dataset from a current research project on smart weed control~\footnote{\url{https://www.photonikforschung.de/projekte/sensorik-und-analytik/projekt/hopla.html} (accessed on April 22, 2024)} (see Acknowledgement~\ref{subsec:ack}), which exhibits distinctive characteristics of weeds and background conditions from experimental settings.

In this paper, we extend our prior work~\cite{stab_diff} by offering an in-depth evaluation of data efficiency through the gradual substitution of real-world images with synthetic counterparts in training lightweight You Only Look Once (YOLO) models ~\cite{Jocher_Ultralytics_YOLO_2023}, intending to deploy  for real-time weed detection tasks. Additionally, we perform an extensive comparison of synthetic versus real-world image quality using various no-reference image quality assessment (NR-IQA) metrics. Furthermore, we discuss a newly implemented automated annotation process incorporated into our pipeline, utilizing the full potential of the YOLOv8x (extra-large) model to effectively annotate synthetic images, thus enhancing our GenAI-based pipeline infrastructure for intelligent weed management systems.


Next to a brief purely qualitative comparison of the generated synthetic images with the real images, we conduct a quantitative image quality assessment as well as a comprehensive task-specific evaluation of the downstream task of weed detection in Section~\ref{sec:eval}. 
For task-specific evaluation, we focused on data efficiency by targeting edge devices with less computational power for real-time weed detection tasks. Consequently, we tested the \textit{nano} and \textit{small} variants of the latest YOLO models to evaluate their efficiency by integrating synthetic images into the training dataset.
In Section~\ref{sec:Conclusion_Future_Work} we discuss our findings and briefly touch upon how our approach can be utilized to enhance the self-improvement capabilities of Organic Computing and other intelligent technical systems by introducing self-reflection. Finally, the paper is closed with an outlook on future research.

The contributions of this paper are summarized as follows:
\begin{enumerate}
    \item \textbf{Novel image data augmentation methodology:} We show how to combine the segment anything foundation model (SAM) with a fine-tuned Stable Diffusion Model within a dedicated data processing pipeline. This allows a) automatic transformation of an object detection dataset into an instance segmentation dataset to subsequently mask the relevant object classes and thereby get rid of the complex background while preserving the exact image shape, and, b) further use of these masks for training a text-guided image generation method, allowing to synthesize specific and tailored artificial training data.
    \item \textbf{Increase data efficiency:} Thereby, our method facilitates increased exploitation of the available training data (next to only feeding to the training process) and is expected to allow for alleviation of the vast training data requirements of deep learning for obtaining robust models. 
    \item \textbf{Increase autonomous learning ability:} Integration of our method into intelligent technical systems architectures potentially serves as one way to increase their autonomous learning and self-improvement capabilities by including synthetic image generation and training into a continual self-reflection and learning loop. 
\end{enumerate}

\section{Background}
\label{sec:Background}
This section 
Provides a brief overview of image augmentation (Sect.~\ref{subsec:image_aug}), the Segment Anything Model (SAM) (Sect.~\ref{subsec:SAM}), diffusion models (Sect.~\ref{subsec:Diff}), as well as the prominent YOLO models for object detection tasks (Sect.~\ref{subsec:yolo}).

\subsection{Image Augmentation}
\label{subsec:image_aug}

Data augmentation methods in computer vision are categorized into model-free and model-based approaches~\cite{xu2023comprehensive}. Despite their effectiveness in downstream tasks, model-free techniques, such as blurring, occlusion, noise addition, and viewpoint changes, lack fidelity and natural variation~\cite{xu2023comprehensive}. In contrast, model-based methods, employing generative AI such as GANs~\cite{GANsNIPS2014_5ca3e9b1}, VAEs~\cite{kingma2013auto}, and Diffusion Models (DMs)~\cite{sohl2015deep} offer greater natural variations and fidelity.
Among these generative AI approaches, DMs showed superior image quality in comparison with the VAEs and GANs~\cite{yang2023diffusion}. Despite their efficacy, DMs remain less utilized in agriculture. Recent studies however demonstrate their effectiveness in weed dataset augmentation~\cite{chen2024synthetic}. Moreover, diffusion models have been found effective for augmenting plant disease datasets~\cite{muhammad2023harnessing}. Furthermore, a recent study found that using synthetic images generated by the Stable Diffusion Model for image augmentation improved YOLO model performance in weed detection compared to traditional augmentation methods~\cite{Modak2024WeedDetectionGenAI}.


\subsection{Segment Anything Model (SAM)}
\label{subsec:SAM}
SAM is a ``zero-shot transfer'' method, which generally allows the segmentation of any object without the need for additional training or fine-tuning. It is trained on one billion image masks and 11 million images. SAM consists of three components, i.e., an image encoder, a prompt encoder, and a mask decoder. The image encoder is realized by a Vision Transformer (ViT), pre-trained by a Mask autoencoder (MAE) method.  The prompt encoder has two types of acceptable prompts, \textit{sparse}, e.g., points, boxes, and text, and \textit{dense} e.g., masks.
%
In the inference stage, SAM can operate in a ``Manual Prompting'' and a ``Fully Automatic'' mode. For the former, manually created texts, boxes, or points are provided as a hint (i.e., conditioning) to predict the image masks. In the latter, Fully Automatic mode, SAM predicts the masks from the input image without any further conditioning input. SAM is available in three pre-trained model variants: ViT-B, ViT-L, and ViT-H, with $91M$, $308M$, and $636M$ learning parameters (neural network weights) respectively. ViT-H considerably outperforms the other models ~\cite{kirillov2023segment}. For weed detection in agriculture, SAM has accelerated annotation tasks, such as weed segmentation via bounding box and point inputs~\cite{carraro2023segment}.


\subsection{Diffusion Models (DMs)}
\label{subsec:Diff}
Amongst generative models, diffusion models are at the forefront and capable of producing realistic images, and high-quality data~\cite{Cao10419041}. Inspired by non-equilibrium statistics, the diffusion process involves the use of two Markov chains, namely forward diffusion and reverse diffusion~\cite{sohl2015deep}. In the forward phase~\ref{eq:for_diff}, each input image $x_0$ undergoes iterative transformation using forward transition kernels $F_t$ parameterized by noise levels $\sigma_t$, leading to the formation of a prior distribution $F(x_0, \sigma)$. This process involves the composition of transition kernels over time $t$, resulting in the gradual refinement of the distribution. Conversely, the reverse diffusion phase~\ref{eq:rev_diff}, reverses this transformation by iteratively applying so-called reverse transition kernels $R_t$ in backward order, guided by noise levels $\sigma_t$. This phase aims to reconstruct the original images $x_T$ from the prior distribution, thus completing the reversal of the forward diffusion process.
\begin{equation}
\label{eq:for_diff}
F(x_0, \sigma) = F_T(x_{T-1}, \sigma_T) \circ \ldots \circ F_t(x_{t-1}, \sigma_t) \circ \ldots \circ F_1(x_0, \sigma_1)
\end{equation}
\begin{equation}
\label{eq:rev_diff}
R(x_T, \sigma) = R_1(x_1, \sigma_1) \circ \ldots \circ R_t(x_t, \sigma_t) \circ \ldots \circ R_T(x_T, \sigma_T)
\end{equation}

During inference time, new images are generated by the gradual reconstruction from white random noise, parameterized by a deep neural network~\cite{ho2020denoising}, typically a U-Net~\cite{ronneberger2015u}. In contrast to other diffusion models, so-called \textit{latent diffusion models} (LDMs), as introduced by Rombach et al. (2022)~\cite{rombach2022high}, minimize computational costs and complexity by leveraging the latent space of a pre-trained autoencoder rather than operating within the pixel space. 
The training is divided into two phases: Firstly, an autoencoder is trained to create an efficient, lower-dimensional representation of the data or image space. Unlike previous methods, DMs are trained in this learned latent space, resulting in better scalability concerning spatial dimensions. LDMs thus allow for efficient image generation with a single network pass. 
Particularly, the autoencoding stage needs to be trained only once, enabling its reuse across multiple DM training or its transfer to different tasks. Additionally, this approach is extended to incorporate transformers into the DM's UNet backbone, facilitating various token-based conditioning mechanisms for image-to-image tasks~\cite{rombach2022high}. The so-called \textit{Stable Diffusion Model}, a text-to-image-based  LDM, has been developed by researchers from CompVis\footnote{\url{https://ommer-lab.com/} (accessed on 6 March 2024)}, Stability AI\footnote{\url{https://stability.ai/}(accessed on 6 March 2024)}, and LAION\footnote{\url{https://laion.ai/}(accessed on 6 March 2024)} and was trained on the LAION-5B~\cite{schuhmann2022laion} dataset, the largest freely accessible multi-modal dataset, containing text-image pairs. The output image can be controlled by the prompted text through a classifier-free guidance~\cite{ho2022classifier} mechanism, ensuring \textit{Stable Diffusion} for precise manipulation of desired visual attributes and generation of high-fidelity images. However, large text-to-image models lack the ability to replicate visual characteristics from reference sets and generate diverse interpretations~\cite{ruiz2023dreambooth}. For dealing with this issue, Ruiz et al. introduced Dreambooth~\cite{ruiz2023dreambooth}, a \textit{few-shot} fine-tuning method for personalizing text-to-image models, addressing subject-specific user needs.
This involves embedding a pair of \textit{unique identifiers} and \textit{subject classes}, such as ``\textit{a HoPla Sugarbeet}''\footnote{HoPla is the research project acronym in which scope this work is conducted. Please see the Acknowledgement\ref{subsec:ack}.}, into the text-to-image diffusion model's \textit{dictionary}. Consequently, the model is enabled to learn the specific subject associated with the \textit{unique identifier}, simplifying the need for verbose image descriptions. Even more importantly, utilizing the \textit{unique identifier}, the model can learn to mimic the style of the input images.
Thus, during the inference stage, an image can be produced by a descriptive text prompt, such as ``a [\textit{unique identifier}] [\textit{subject classes}] [\textit{context description}]''. 
Using this unique approach, the subject can be placed in different backgrounds with realistic integration, including e.g., shadows and reflections.
As we describe in the subsequent Section~\ref{sec:Current_Approach}, we also make use of these various methods and integrate them into a pipeline architecture for the generation of synthetic high-fidelity weed images for data-efficient training of robust weed detection models.


\subsection{You Only Look Once (YOLO) Models}
\label{subsec:yolo}
In the domain of real-time object detection, YOLO models have gained significant traction due to their impressive speed and efficiency. YOLO approaches the object detection task as a regression problem by dividing the input image into an $S \times S$ grid and predicting bounding boxes ($B$) and class probabilities ($C$) in a single pass. Each prediction consists of five regression values: $P_c$ (confidence score), $b_x$ and $b_y$ (center coordinates of the bounding box), and $b_h$ and $b_w$ (dimensions of the bounding box). The output is an $S \times S \times (B \times 5 + C)$ tensor, which is further refined using non-maximum suppression (NMS) to eliminate duplicate detections~\cite{Jocher_Ultralytics_YOLO_2023}.

YOLO models are evaluated using commonly used metrics for assessing the performance of object detection models, such as \textit{precision, recall, F1 score, mAP50}, and\textit{ mAP50\text{-}95}. \textit{Precision} measures the proportion of correctly identified positive instances (see eq.~\ref{eq:precision}); in the context of weed detection, high precision means that when the model identifies a weed, it is likely correct. 
\begin{equation}
\text{Precision} = \frac{\text{True Positives}}{\text{True Positives} + \text{False Positives}}
\label{eq:precision}
\end{equation}
\textit{Recall} measures the proportion of actual positives that are correctly detected (see eq.~\ref{eq:recall}), in case of weed detection indicating the model's ability to identify all present weeds. Moreover, a high recall value in weed detection indicates that the model can effectively detect high proportions of actual weeds from the dataset~\cite{liu2023automated}.
\begin{equation}
\text{Recall} = \frac{\text{True Positives}}{\text{True Positives} + \text{False Negatives}}
\label{eq:recall}
\end{equation}
The \textit{F1 score} is calculated to provide a single metric that balances both precision and recall, representing the overall effectiveness of the detection model (see eq.~\ref{eq:F1}).
\begin{equation}
F1 = 2 \cdot \frac{\text{Precision} \cdot \text{Recall}}{\text{Precision} + \text{Recall}}
\label{eq:F1}
\end{equation}
\textit{Intersection over Union (IoU)} is an essential metric for object localization in detection tasks, quantifying the overlap between predicted and ground truth bounding boxes. Average Precision (AP) computes the area under the precision-recall curve, providing an overall measure of the model's performance, while Mean Average Precision (mAP) extends this concept by averaging the precision across all object classes. \textit{mAP50} calculates mAP at an IoU threshold of 0.50, while \textit{mAP50\text{-}95} computes mAP across various IoU thresholds from 0.50 to 0.95. For comprehensive performance evaluation with reduced localization error,\textit{ mAP50\text{-}95} is typically preferred.

 YOLO models have been employed in a wide range of fields, including autonomous driving, medical applications, and autonomous weed, crop, and pest detection~\cite{terven2023comprehensive}. The latest models in the YOLO series, such as YOLOv8, YOLOv9, and YOLOv10 have further expanded the applicability and versatility of the approach in numerous domains in real-time detection~\cite{Jocher_Ultralytics_YOLO_2023}. The YOLO series comprises various models including \textit{nano, small, medium, large}, and \textit{extra large}, tailored to different hardware capabilities. Due to the limited resources of edge devices such as the Raspberry Pi\footnote{\url{https://www.raspberrypi.com/for-industry/} (accessed on: 04 July 2024)} and NVIDIA Jetson\footnote{\url{https://www.nvidia.com/de-de/autonomous-machines/embedded-systems/} (accessed on: 04 July 2024)}, DL models with a reduced number of parameters are often chosen for these platforms. Thus, for deployment on edge devices, the \textit{nano} model is selected for highly resource-limited settings, and the 	\textit{small} YOLO model is favored to maintain a balance between speed and accuracy. In terms of latency and mAP50\text{-}95 score, the YOLOv10 nano and small variants outperformed the corresponding variants of other state-of-the-art YOLO models (cf. Tab.~\ref{tab:yolo}). 

\begin{table}[h!]

\centering
\caption{Comparison of YOLOv8, YOLOv9, and YOLOv10 nano and small models based on parameters, latency, and mAP50\text{-}95 on the COCO dataset (image size: 640 pixels, TensorRT FP16 on T4 GPU)~\cite{Jocher_Ultralytics_YOLO_2023}.}

\begin{tabular}{lcccccc}
\toprule
\textbf{Model} & \multicolumn{2}{c}{\textbf{Parameters (M)}} & \multicolumn{2}{c}{\textbf{Latency (ms)}} & \multicolumn{2}{c}{\textbf{mAP50\text{-}95}} \\
\cmidrule(r){2-3} \cmidrule(r){4-5} \cmidrule(r){6-7}
 & \textbf{Nano} & \textbf{Small} & \textbf{Nano} & \textbf{Small} & \textbf{Nano} & \textbf{Small} \\
\midrule
\textbf{YOLOv10} & 2.3  & 7.2  & 1.8  & 2.49  & 39.5 & 46.8 \\
\textbf{YOLOv9}  & 2.0  & 	7.2  & --   & --  & 38.3 & 46.8 \\
\textbf{YOLOv8}  & 3.2  & 11.2  & 6.16 & 7.07  & 37.3 & 44.9 \\
\bottomrule
\end{tabular}
\label{tab:yolo}
\end{table}

Different versions of YOLO nano and its modified variants have demonstrated their potential in various agricultural use cases, such as detection of color changes in ripening melons~\cite{chen2024yolov8}, real-time apple fruit detection~\cite{ma2024using}, monitoring the stages of cabbage head harvest stages~\cite{tian2024recognition}, detecting small strawberries~\cite{luo2024small}, detection of weeds in sugar beet fields~\cite{Saltik2024WeedDetection},~\cite{Modak2024WeedDetectionGenAI}.

\section{Methodological Approach}
\label{sec:Current_Approach}
\begin{figure}[!htb]
\centering\includegraphics[width=1.0\columnwidth]{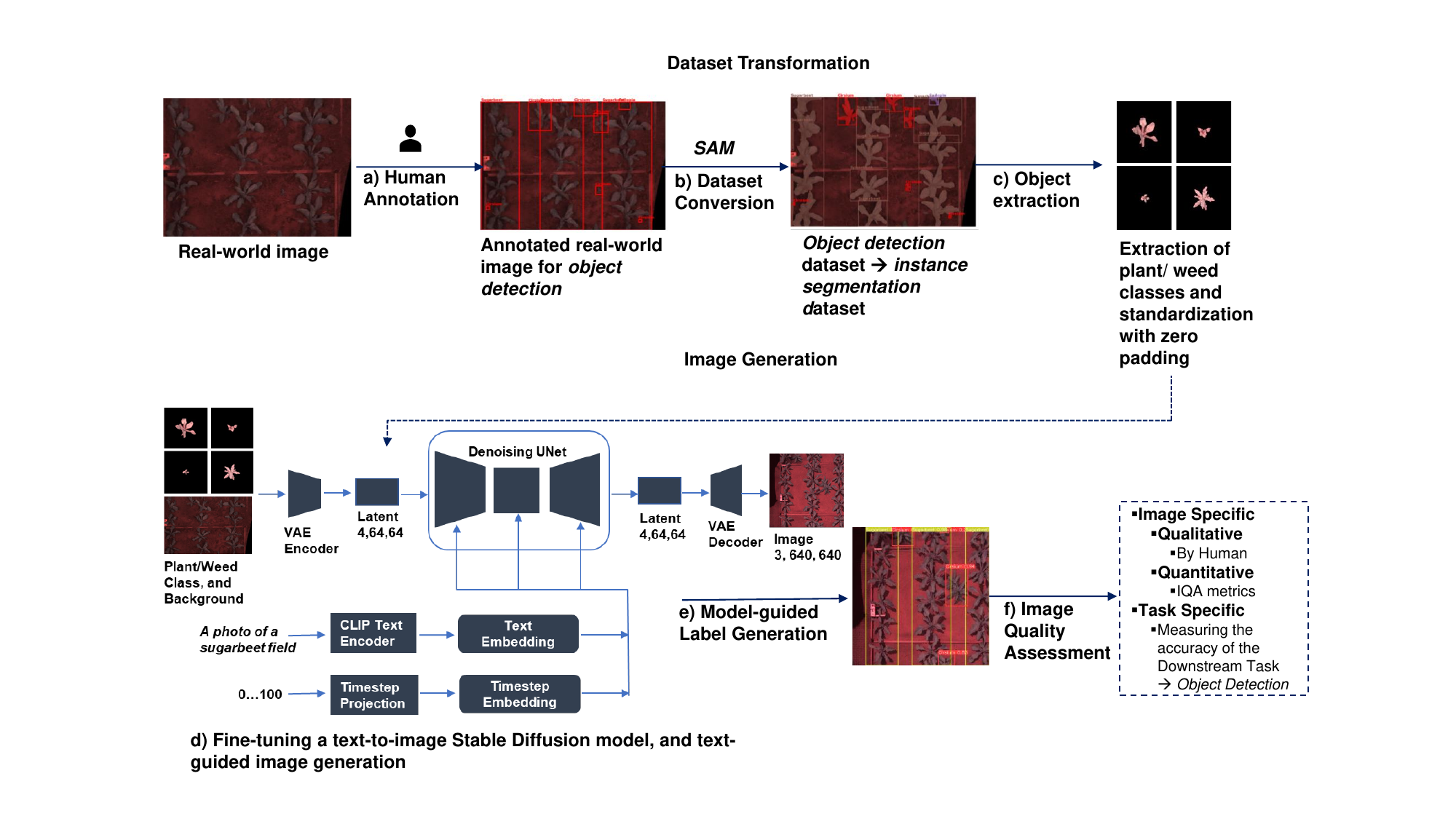}
\caption{Synthetic object (here weed) generation pipeline. The upper half shows the \textit{dataset Transformation} phase: Utilization of the foundation model SAM to convert object detection datasets into instance segmentation datasets (b). Weed classes are masked to eliminate complex backgrounds while preserving image integrity (c). \textit{Image Generation} phase (lower half): Fine-tuning of a \textit{Stable Diffusion Model} using weed masks and background images to facilitate text-guided image generation (d), and subsequent model-guided label generation (e). The last step (f) shows options to perform image quality assessment (IQA), such as \textit{image specific} methods, including quantitative and qualitative metrics, and \textit{task specific} approaches prescribed by the downstream task, i.e., mAP score for object detection}
\label{fig:pipe}
\end{figure}

We consider the agricultural use case of weed detection -- a prerequisite for every smart spraying or autonomous weed control system (e.g., weeding robots). 
Accordingly, an image processing pipeline for generating synthetic images of weed-infested soil areas is developed, which is depicted in Figure~\ref{fig:pipe}.
Our approach integrates the unique capabilities of the \textit{Segment Anything Model} (SAM) and a generative \textit{Stable Diffusion Model}. 
We proceed by describing the dataset from a current research project, the sensor used to collect it, as well as the resulting data modality. 
Subsequently, we detail the first phase of our synthetic image generation pipeline, which we call the \textit{ data transformation phase}.
One crucial step in this phase is to leverage the universal segmentation feature of the foundation model SAM when applied to pre-annotated training data. For this initial paper, we assumed that the collected training data needs to be human-annotated. 
The subsequent image generation phase employing the \textit{Stable Diffusion Model}  is described afterward. After generating synthetic images, we employed a fine-tuned \textit{YOLOv8x} model for label prediction and annotating these images. Later, the synthetic images were evaluated using no-reference IQA metrics against real-world images. In addition, we used a strategic substitution of real-world images with synthetic ones to evaluate the data efficiency in training the \textit{nano} and \textit{small} versions of the \textit{YOLOv8, YOLOv9, and YOLOv10} models.

We utilized an NVIDIA A100-SXM4-40GB GPU with 40 GB of VRAM and allocated 4 CPU cores and 12 GB of system memory on an AMD EPYC 75F3 32-core Processor for all stages of the Stable Diffusion training, image, and label generation, as well as YOLO models training and evaluation.

\subsection{Dataset}
\label{subsec:data}
\begin{figure}[!htb]
  \centering
  \begin{subfigure}[b]{0.49\textwidth}
    \centering
    \includegraphics[width=\textwidth]{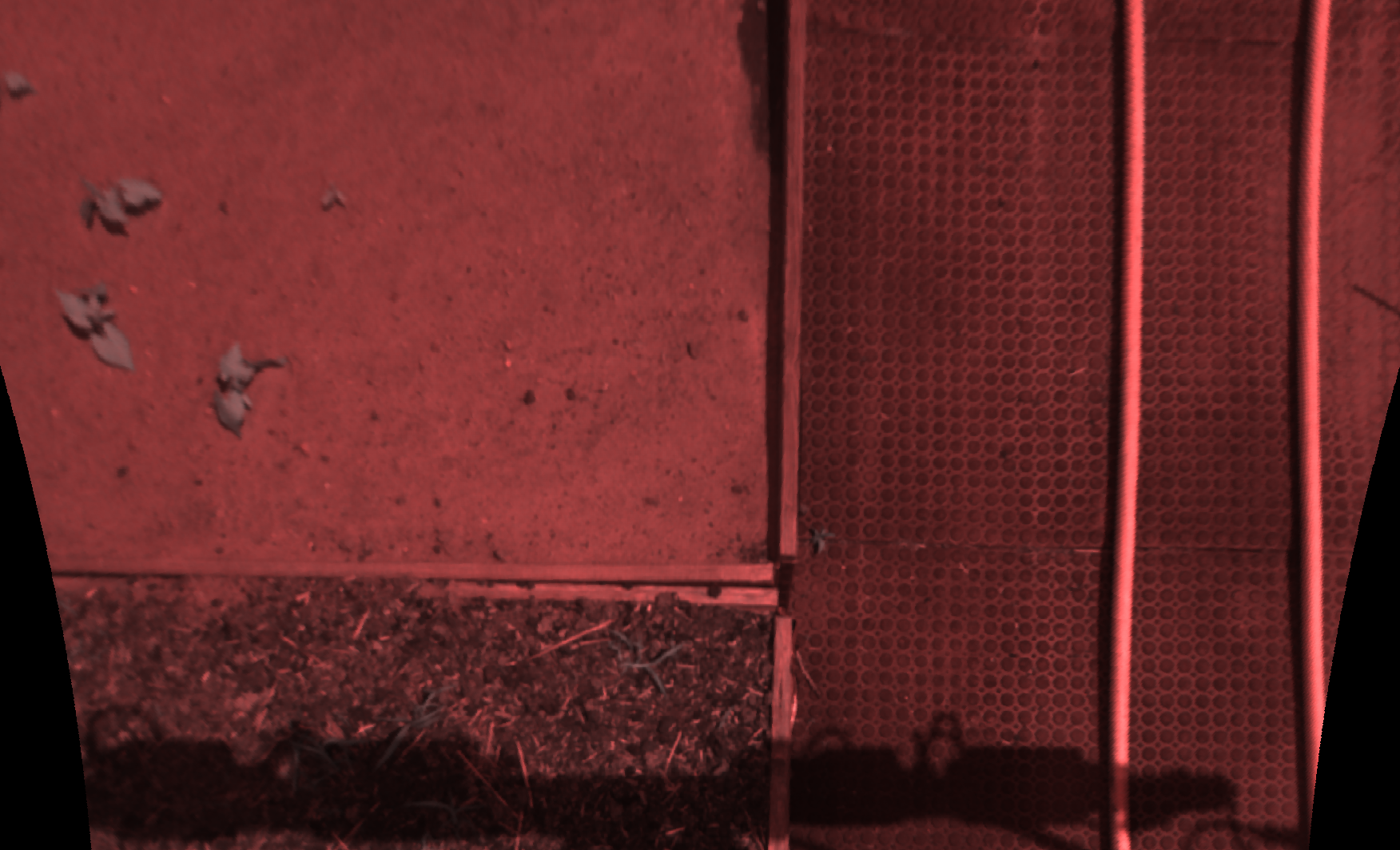}
    \label{fig:img3}
  \end{subfigure}\hfill
  \begin{subfigure}[b]{0.49\textwidth}
    \centering
    \includegraphics[width=\textwidth]{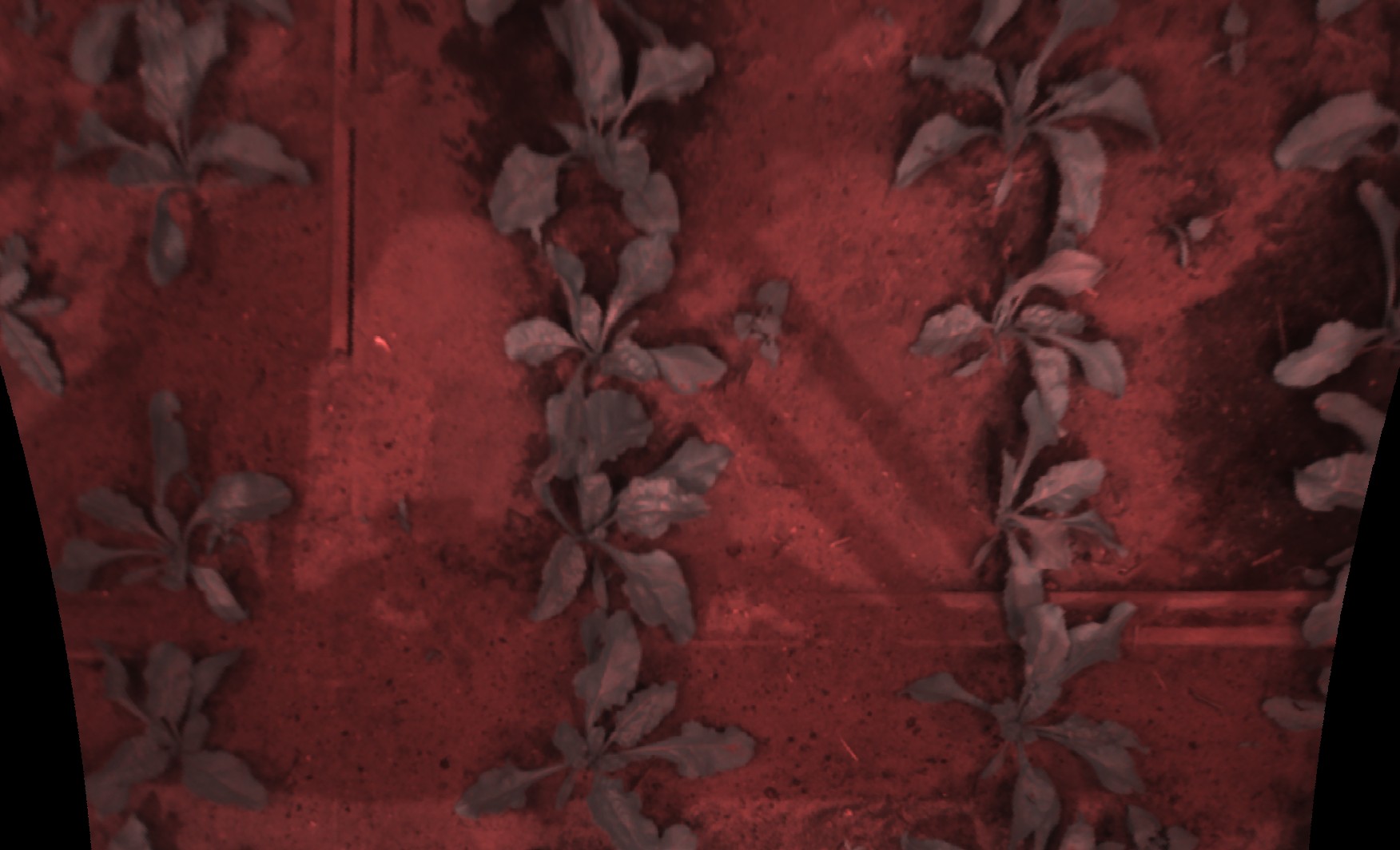}
    \label{fig:img4}
  \end{subfigure}
  \caption{ Sample pseudo-RGB images from the sugar beet dataset, captured from the described outdoor euro pallet setup. The dataset comprises \textit{sugar beet} and weed classes (\textit{Cirsium, Convolvulus, Fallopia}, and \textit{Echinochloa})
  }
  \label{fig:real}
\end{figure}

The dataset was collected in the scope of a current research project (see Acknowledgement \ref{subsec:ack}) at an experimental site in Rhineland Palatinate, Germany.
The industrial camera sensor was attached to a herbicide sprayer mounted on a tractor which was operating at a speed of \SI{1.5}{\metre\per\second}.
Within the field camera unit (FCU), a 2.3-megapixel RGB camera sensor with an Effective Focal Length (EFL) of 6mm was employed. 
This camera sensor was equipped with a dual-band filter tailored for capturing RED and near-infrared (NIR) wavelengths. 
The multiple FCUs attached at the sprayer's linkages maintained a constant height of 1.1 meters above the ground at a 25-degree off-vertical angle.
The land machine carrying the camera sensors moves along a controlled, outdoor experimental setup where different crops and weeds are grown under various soil conditions in boxes built on euro pallets marked accordingly to identify the different weed and soil types. 
This method has been chosen to obtain well-balanced datasets for training robust weed detectors. 
Subsequently, pseudo-RGB images were derived from the raw RED and NIR bands after performing projection correction. The dataset was then manually labeled by domain expert annotators.
The dataset comprises 2074 images primarily featuring sugar beet as the main crop class, alongside four weed classes: \textit{Cirsium}, \textit{Convolvulus}, \textit{Fallopia}, and \textit{Echinochloa}. The images have a resolution of \(1752 \times 1064\) pixels. As exemplarily shown in Figure~\ref{fig:real}, our dataset exhibits distorted and unusual features. Most prominently the images are characterized by showing misaligned pallet frames, off-centered boxes, blurred pixels, and very small weeds. Despite that, we chose to utilize this dataset for our experiments due to the following reason: Such idiosyncratic features are commonly encountered in real-world settings and our objective is to demonstrate the robustness of our initial method regarding its general ability to capture and generate such idiosyncrasies effectively. In future research, we will extend our experiments by using several datasets.

\subsection{Dataset Transformation} 

The original dataset was initially annotated for object detection tasks (step a) in Fig.~\ref{fig:pipe}). However, to refine the delineation of plants and weeds and eliminate unwanted backgrounds, we transformed bounding boxes into complex polygons using the \textit{Segment Anything Model} (SAM). This conversion effectively converted the object detection dataset into an instance segmentation dataset, i.e., containing segmentation masks of the individual plants within an image.  Given the complexity of the agricultural dataset and SAM's tendency for over-segmentation, achieving ``Fully Automatic'' image segmentation was not feasible with the current version of SAM~\cite{carraro2023segment}. Consequently, we have applied a custom prompting method on SAM $ViT-H$ to semi-automate the dataset conversion process by employing batch bounding box prompting (cf. step b) in Figure\ref{fig:pipe})~\cite{kirillov2023segment}. Subsequently, in our pipeline's step c), a custom Python script is added which masks the plant and weed classes. To address variations in mask sizes, we applied standardization by resizing the masks to \(512 \times 512\) pixels with zero padding which in turn avoids up-scaling and thus image deformation~\cite{hashemi2019enlarging}.

\subsection{Image Generation}
The image generation phase of the proposed pipeline comprises two main steps: (1) fine-tuning a\textit{ Stable Diffusion Model} and (2) text-to-image inference (cf. step d) in Fig.~\ref{fig:pipe}).
Pursuing the creation of a diverse synthetic dataset, while still preserving important subject features (plants \& weeds shapes), we choose subject-specific training to fine-tune the \textit{Stable Diffusion $1.5$ model}, employing the previously mentioned technique \textit{Dreambooth} (cf. Sect.~\ref{subsec:Diff}). 
This involves training the \textit{Stable Diffusion Model} with embeddings representing ``HoPla,'' an \textit{Unique Identifier}, and various \textit{Subject Classes}, including \textit{Sugar beet}, \textit{Cirsium}, \textit{Echinchloa}, \textit{Fallopia}, \textit{Convolvulus}, and the background \textit{plots}. An example prompt would look like this: \textit{``A Photo of HoPla Sugar beet with HoPla Fallopia on HoPla Plot''}. Although Dreambooth was originally designed to train one subject at a time, our dataset comprises five distinct plant/weed classes and a background soil class. Leveraging the \textit{diffusers} library~\cite{von-platen-etal-2022-diffusers}, we implemented a process called multi-subject Dreambooth\footnote{\url{https://github.com/huggingface/diffusers/tree/main/examples/research_projects/multi_subject_dreambooth/} (accessed on 6 March 2024)}
to simultaneously train the model on multiple subject classes. 
We performed the fine-tuning of the \textit{Stable Diffusion Model} by varying the model conditioning prompts, with a maximum of 60000 training steps. Additionally, we employed a batch size of 1 and utilized gradient checkpointing and accumulation techniques to enhance memory efficiency. Furthermore, we trained a text encoder to incorporate the \textit{Unique identifier} and \textit{Subject classes} into the diffusion model.

After completing the fine-tuning of the generative model by the abovementioned method, the inference stage offers controllability and customization regarding particular image generation through explicit prompting. This allows the optimization of different parameters, such as \textit{text prompt description}, \textit{number of denoising steps}, \textit{image resolutions}, and \textit{guidance scale}  according to the user's needs. In this stage, we utilized the diffusion pipeline by making use of the diffusers library~\cite{von-platen-etal-2022-diffusers} once again. In our case, we generated images by optimizing various \textit{text prompts}, \textit{number of inference steps}, and \textit{guidance scale} variants to assess image generation quality, inference time, and the alignment of the text prompt with the generated image, respectively, according to our requirements. The text prompt represents a typical description of images, while the \textit{number of inference steps} indicates the number of iterative diffusion steps required to generate an image from the noise tensor. Generally, a higher number of inference steps leads to higher quality and better-detailed images, as it involves more iterations to reduce noise and enhance details~\cite{von-platen-etal-2022-diffusers}. The choice of the number of inference steps depends on factors such as desired image quality, available computational resources, and specific dataset characteristics. Finding the optimal number of inference steps depends on the practitioner's requirements and the computational resources available. In our research, we found that using $50$ inference steps best suited our needs and hardware setup. The \textit{guidance scale} determines the linkage of generated images to the text prompt. The default \textit{guidance scale} is $7.5$, however, the higher the value of \textit{guidance scale}, the higher the generated images associated with the \textit{text prompt}~\cite{von-platen-etal-2022-diffusers}. In our experiment, we kept the default setting of the \textit{guidance scale}. Moreover, the quality and characteristics of generated images can be controlled by different schedulers. These schedulers control how denoising steps progress over time, influencing image quality and fidelity. As with other parameters in the pipeline, practitioners can independently experiment with various schedulers based on their needs, as there are no quantitative metrics available for evaluation. In our use cases, we have observed that the Euler Ancestral Discrete Scheduler\footnote{\url{https://huggingface.co/docs/diffusers/en/api/schedulers/euler_ancestral} (accessed on 6 March 2024)} consistently generated the desired quality of images in $20-30$ denoising steps.

\subsection{Label Generation} During the \textit{image generation} phase (see Fig.~\ref{fig:pipe}(step d)), we initially categorized the weeds into four distinct species: Cirsium, Convolvulus, Fallopia, and Echinochloa. However, in the subsequent \textit{label generation} and \textit{weed detection} phase, we reclassified these species into two broader botanical categories to enhance practical applicability: dicotyledons (Cirsium, Convolvulus, and Fallopia) and monocotyledons (Echinochloa). This reclassification was performed to better align with the herbicides available on the market, which are designed to target specific botanical categories rather than individual species~\cite{herrera2014novel}.

We fine-tuned a pre-trained YOLOV8x model, originally trained on the COCO dataset~\cite{DBLP:journals/corr/LinMBHPRDZ14}, using our real-world sugar beet dataset with three classes: Sugar beet, Monocotyledons (Monocot), and Dicotyledons (Dicot). The fine-tuned weight from the YOLOV8x model was utilized to annotate our synthetic images through an automated, model-guided annotation technique. This enhancement further optimizes our GenAI-based pipeline architecture for intelligent weed control systems, substantially reducing the time, cost, and labor associated with the annotation process.


\subsection{Evaluation}

\label{sec:eval}
\begin{figure}[!htb]
  \centering
  
  \begin{subfigure}[b]{0.24\textwidth}
    \centering
    \includegraphics[width=\textwidth]{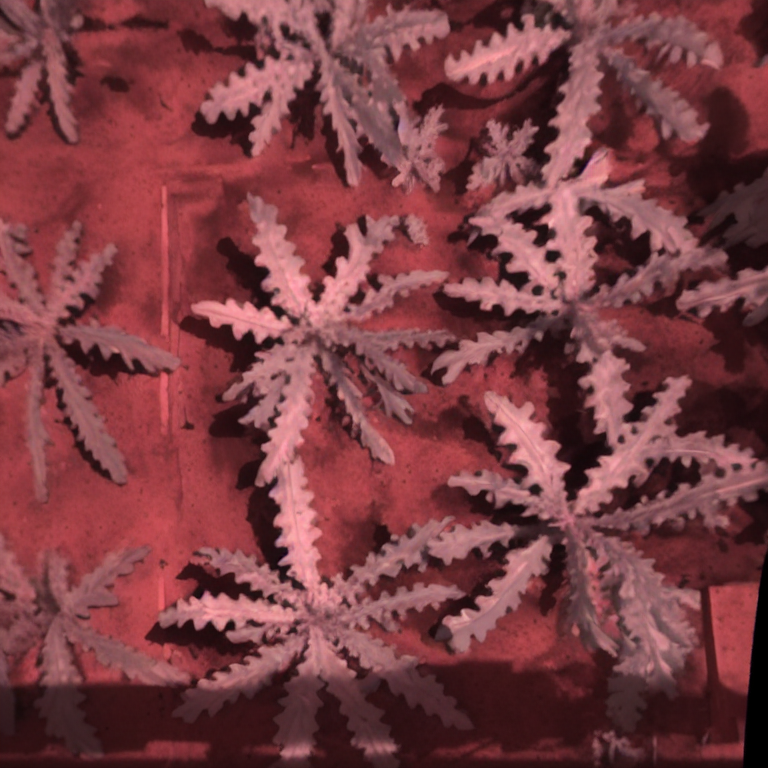}
    \caption{Cirsium}
    \label{fig:cir}
  \end{subfigure}\hfill
  \begin{subfigure}[b]{0.24\textwidth}
    \centering
    \includegraphics[width=\textwidth]{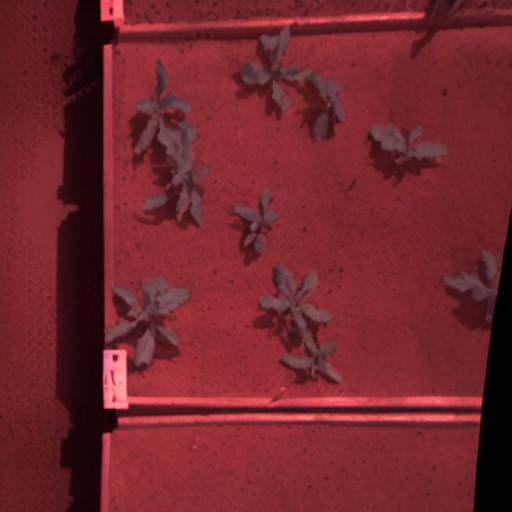}
    \caption{Convolvulus}
    \label{fig:conv}
  \end{subfigure}\hfill
  \begin{subfigure}[b]{0.24\textwidth}
    \centering
    \includegraphics[width=\textwidth]{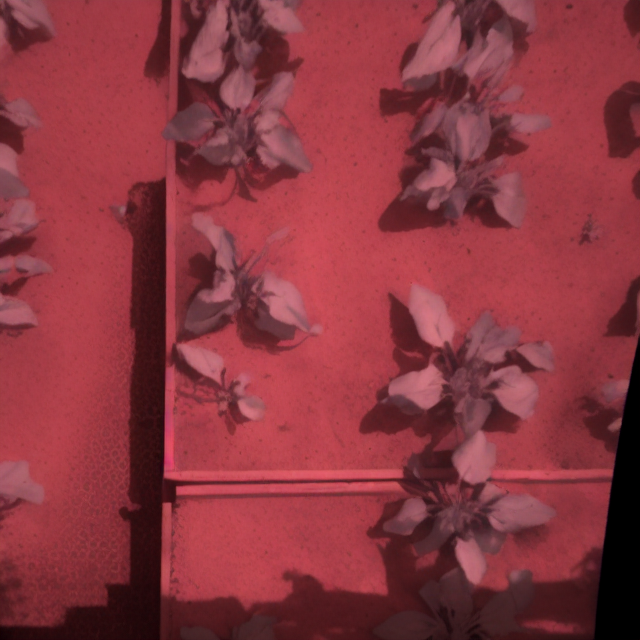}
    \caption{Fallopia}
    \label{fig:fall}
  \end{subfigure}\hfill
  \begin{subfigure}[b]{0.24\textwidth}
    \centering
    \includegraphics[width=\textwidth]{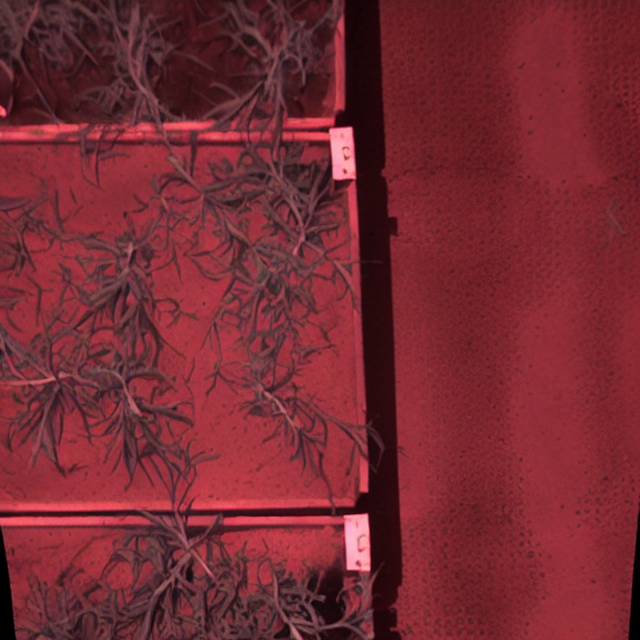}
    \caption{Echinochloa}
    \label{fig:Echi}
  \end{subfigure}
  
  \caption{Synthetic images generated by \textit{fixed weed class} mode, depicting different weed classes: (a) \textit{Cirsium}, (b)\textit{Convolvulus}, (c)\textit{Fallopia}, (d)\textit{Echinochloa}}
  \label{fig:synthetic_images}
\end{figure}

\begin{figure}[!htb]
  \centering
  
  \begin{subfigure}[b]{0.24\textwidth}
    \centering
    \includegraphics[width=\textwidth]{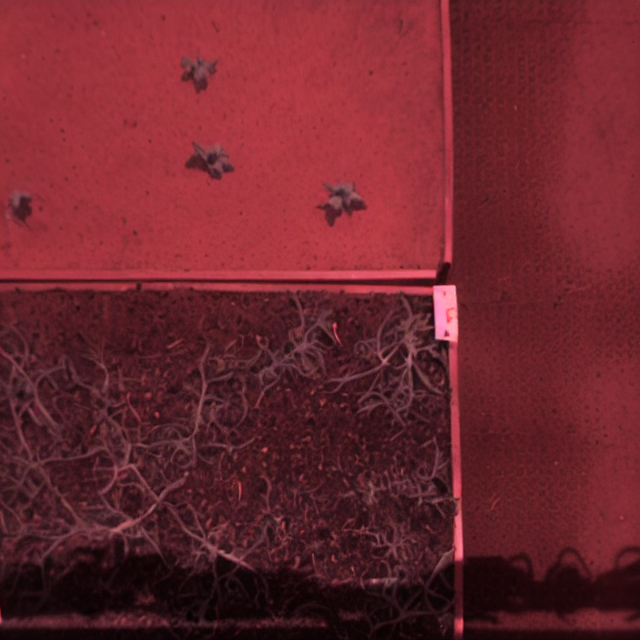}
    
  \end{subfigure}\hfill
  \begin{subfigure}[b]{0.24\textwidth}
    \centering
    \includegraphics[width=\textwidth]{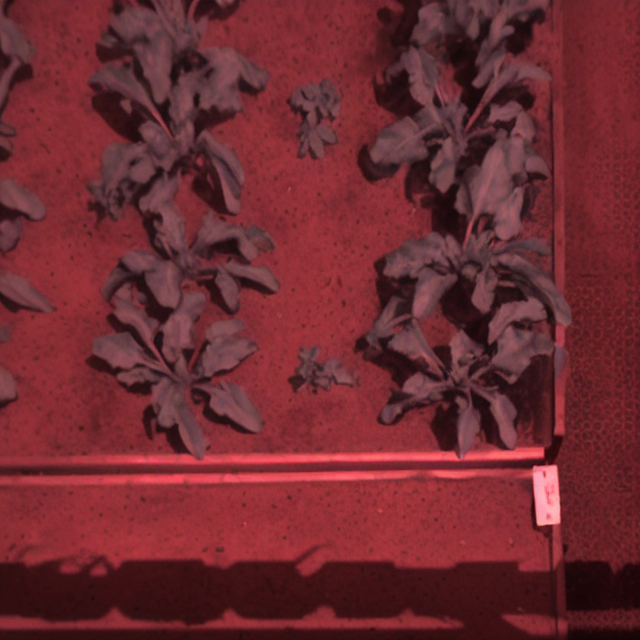}
    
  \end{subfigure}\hfill
  \begin{subfigure}[b]{0.24\textwidth}
    \centering
    \includegraphics[width=\textwidth]{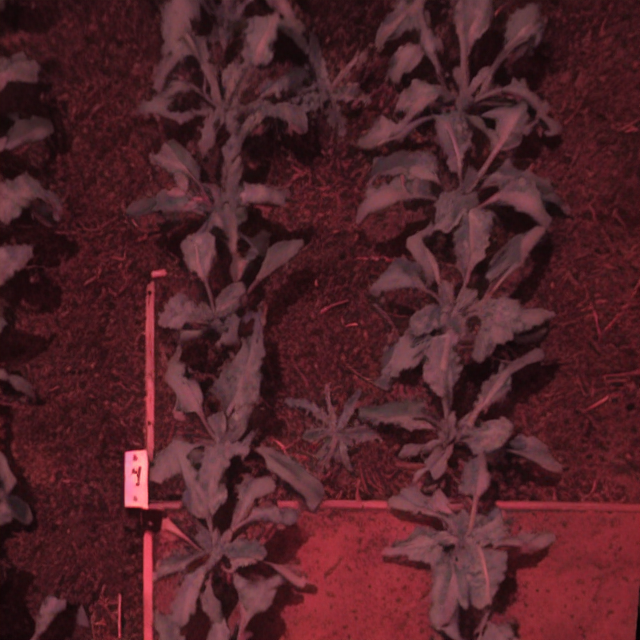}
    
  \end{subfigure}\hfill
  \begin{subfigure}[b]{0.24\textwidth}
    \centering
    \includegraphics[width=\textwidth]{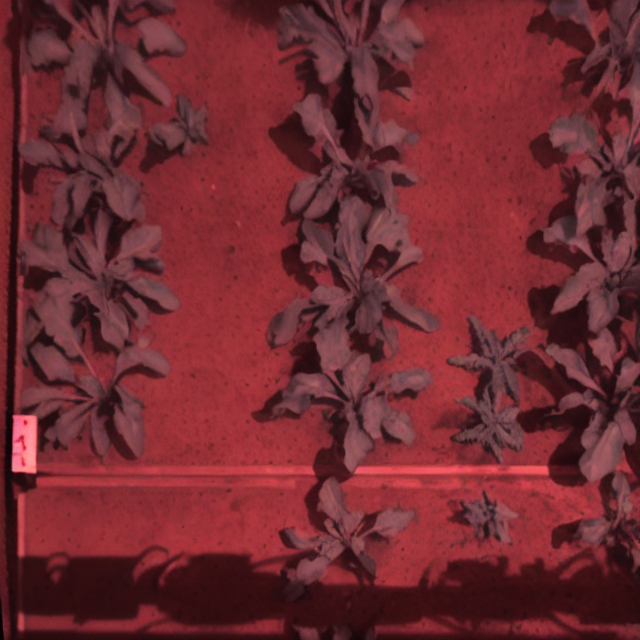}

  \end{subfigure}
  
  \caption{Synthetic images created by \textit{random generation} mode, depicting diversified plant and weed classes on the plots}
  \label{fig:synthetic_images_2}
\end{figure}

The pursued goal of generating synthetic images is to augment the training database and to test the data-efficient training by replacing real-world images with synthetic images for a certain downstream task; in our case, object detection in an agricultural weed control setting. Due to its numerously demonstrated superior detection accuracy and precision, the state-of-the-art object detection model YOLO models~\cite{Jocher_Ultralytics_YOLO_2023} and its variants have been utilized in our study. Since YOLO models operate with a resolution of $640\times640$, we set the resolution of the synthetic images at $640\times640$.
Since Stable Diffusion is a text-to-image model architecture, we used various \textit{text prompts} to evaluate the weed diversity, fidelity, and relation to the real environment. To be able to address common data issues such as `class imbalance' and `lack of diversity', we split our image generation goal into two modes: \textit{fixed weed class} and \textit{random generation}. For instance, we prompted our model for the former \textit{fixed weed class} with prompts such as \textit{`A Photo of HoPla Echinochloa, HoPla Plot in the Background'}, and for the latter \textit{random generation} case with \textit{`A photo of random plants and weeds, HoPla Plot in the Background'}. Samples of synthetic images from the \textit{fixed weed class}, and for \textit{random generation} are depicted in Figures~\ref{fig:synthetic_images}, and~\ref{fig:synthetic_images_2}, respectively.

When compared with the real weed images (cf. Figure~\ref{fig:real}), it can be seen that the generated synthetic images accurately mimic the style of the corresponding real-world settings. Moreover, the images retain the distinctive features of their subject classes and convincingly mimic shadows found in real-world scenes, thus, significantly enhancing the realism of the synthetic images. Most commonly, image quality evaluation is conducted in two specific ways: image-specific and task-specific methods~\cite{modak2024pansharpening}.
Image-specific quality evaluation can be performed qualitatively and quantitatively. The qualitative method relies on human evaluation, which appears as one of the most intuitive metrics for assessing synthetic images. We will employ a technique known as 2-alternative forced-choice (2AFC)~\cite{Thurstone1927}, where two images are presented simultaneously to the human observer, who must choose between the available options: \textit{synthetic} or \textit{real}.

For quantitative evaluation, image quality assessment (IQA) metrics will be used to measure the image spatial quality in terms of, e.g., blurriness, sharpness, and noises in the synthetic images to compare with the resulting values of real images. We have identified  Blind/Referenceless Image Spatial Quality Evaluator (BRISQUE)~\cite{mittal2012no}, Naturalness Image Quality Evaluator (NIQE)~\cite{mittal2012making}, Deep Bilinear Convolutional Neural Network (DBCNN)~\cite{dbcnn}, HyperIQA~\cite{su2020blindly}, and CLIP-IQA~\cite{wang2023exploring} metrics as suitable candidates for quantitative evaluation. These metrics were selected to ensure broad coverage of various image quality dimensions, combining traditional statistical-based approaches (BRISQUE, NIQE) with more advanced deep learning-based techniques (DBCNN, HyperIQA, CLIP-IQA). 
Fréchet Inception Distance (FID)~\cite{heusel2017gans} is another metric that measures the similarity between feature representations extracted from real and images generated by generative adversarial networks (GANs), however, its adaptability to diffusion models has been not confirmed yet. 
BRISQUE is a no-reference image quality assessment metric that relies on natural scene statistics (NSS) to evaluate the degradation in ``naturalness" caused by distortions and assess overall image quality. Additionally, its low computational demands make it ideal for real-time applications. The score ranges from 0 to 100, with lower scores representing higher image quality~\cite{mittal2012no}. Similar to BRISQUE, NIQE is also a general-purpose metric for no-reference image quality assessment. It is based on statistical features aware of quality in the spatial domain of an NSS model. The scale spans from 0 to infinity, where lower values signify superior image quality~\cite{mittal2012making}. DBCNN is a bilinear blind IQA metric constructed from two CNNs. One network is trained on real distortions, while the other is trained on synthetic distortions. Subsequently, the outputs of these networks are combined bilinearly into a unified quality representation. This resulting network is then fine-tuned using a target subject-rated database. It assesses \textit{Gaussian blur, white noise, JPEG compression, contrast changes, pink noise, overexposure}, and \textit{underexposure}, with higher scores indicating better image quality~\cite{dbcnn}. 
HyperIQA is a deep neural network-based no-reference IQA metric, designed to assess authentically distorted images using a self-adaptive hyper network architecture. It evaluates various types of distortions, including \textit{out of focus}, \textit{low illumination}, \textit{motion blur}, and \textit{lighting}, where higher scores represent superior image quality~\cite{su2020blindly}.

CLIP-IQA~\cite{wang2023exploring} harnesses the text-image pair property of Contrastive Language-Image Pre-training (CLIP)~\cite{radford2021learning} to accurately assess images akin to human perception. The working mechanism of CLIP-IQA involves quantifying image quality by calculating the cosine similarity between provided images and predefined prompts. These prompts represent various image properties such as \textit{brightness, noisiness, sharpness, complexity, naturalness}, and \textit{realism}.


To explore the statistical significance of the difference between the scores of real and synthetic images, hypothesis testing was carried out. First, the Shapiro-Wilk test was used to check for normal distribution of the quantitative scores of BRISQUE, NIQE, DBCNN, HyperIQA, as well as different image properties (\textit{brightness, noisiness, sharpness, complexity, naturalness}, and \textit{realism}) as assessed by CLIP IQA, across the entire dataset. Since normality could not be confirmed, the non-parametric Mann-Whitney U test was chosen and performed with a significance level of $\alpha = 0.05$.

Turning to the \textit{task-specific} evaluation of image quality, a key objective of generating synthetic images is to improve the \textit{data efficiency} of the model training process. \textit{Data efficiency} in the context of our study refers to the amount of synthetic training data used to replace real training data while still achieving comparable or even better performance metrics on the downstream task (weed detection) compared to those obtained from training on the same total amount of real-world images only. 
Accordingly, the ultimate goal of our GenAI-based training approach is to reduce the cost and labor associated with collecting and annotating real-world training data while at the same time increasing model robustness through training with data possessing higher diversity. 
To demonstrate the effectiveness of our approach to achieving this goal, we trained the \textit{nano} and \textit{small} variants of the YOLOv8, YOLOv9, and YOLOv10 models for the challenging downstream task of weed detection, with the objective of deploying these models on edge devices.


The performance of the downstream task is evaluated based on the standard mAP50 and mAP50\text{-}95 metrics (see Sect.~\ref{subsec:yolo}). Within the scope of weed detection, mAP50 scores evaluate model accuracy by comparing predicted bounding boxes to ground truth boxes at an IoU threshold of 0.50, providing a comprehensive measure of the model's performance across all weed classes. Meanwhile, mAP50\text{-}95 is calculated as the mean Average Precision (mAP) over IoU thresholds from 0.50 to 0.95, providing a more comprehensive evaluation of the model's weed detection performance across different levels of bounding box localization accuracy. A high mAP50\text{-}95 signifies superior accuracy across all weed categories, without any inherent bias. 



To train YOLO models, we implemented a defined hyperparameter configuration, detailed in Table~\ref{tab:hyp}. The training was carried out over 300 epochs, with early stopping applied to prevent overfitting, utilizing a patience of 30 epochs without improvement. A batch size of 16 was selected to optimize the trade-off between training efficiency and memory consumption. The initial learning rate was set at 0.01, complemented by a cosine learning rate schedule to facilitate dynamic adjustment throughout the training process. In particular, on-the-fly data augmentation was disabled to ensure consistency in the training dataset, enabling a more accurate evaluation of the models' performance across various combinations of real and synthetic datasets.

\begin{table}[H]
\centering
\caption{Hyperparameter configuration for YOLO models training}
\label{tab:hyp}
\begin{tabular}{@{}ll@{}}
\toprule
\textbf{Hyperparameter} & \textbf{Value} \\ \hline
Epochs & 300 \\
Early stopping patience &   30  \\
Batch size & 16 \\
Initial learning rate & 0.01 \\
Learning rate schedule & Cosine  \\
Augmentation & Disabled \\
\bottomrule
\end{tabular}
\end{table}

The dataset was split into training, validation, and test sets with proportions of 70\%, 15\%, and 15\%, respectively. The training set contained \( n_{\text{training}} = 1508 \) samples. To evaluate the effect of synthetic images, we replaced a proportion \( p \) (ranging from 10\% to 90\%) of the real-world images in the training set with synthetic images. The number of synthetic images \( n_{\text{synthetic}} \) was given by:

\[
n_{\text{synthetic}} = p \times n_{\text{training}}
\]

This approach varied \( n_{\text{synthetic}} \) from 10\% to 90\% of the training set size, while keeping \( n_{\text{training}} \) constant at 1508.

\begin{figure}[H]
    \centering
    \includegraphics[width=1.0\textwidth]{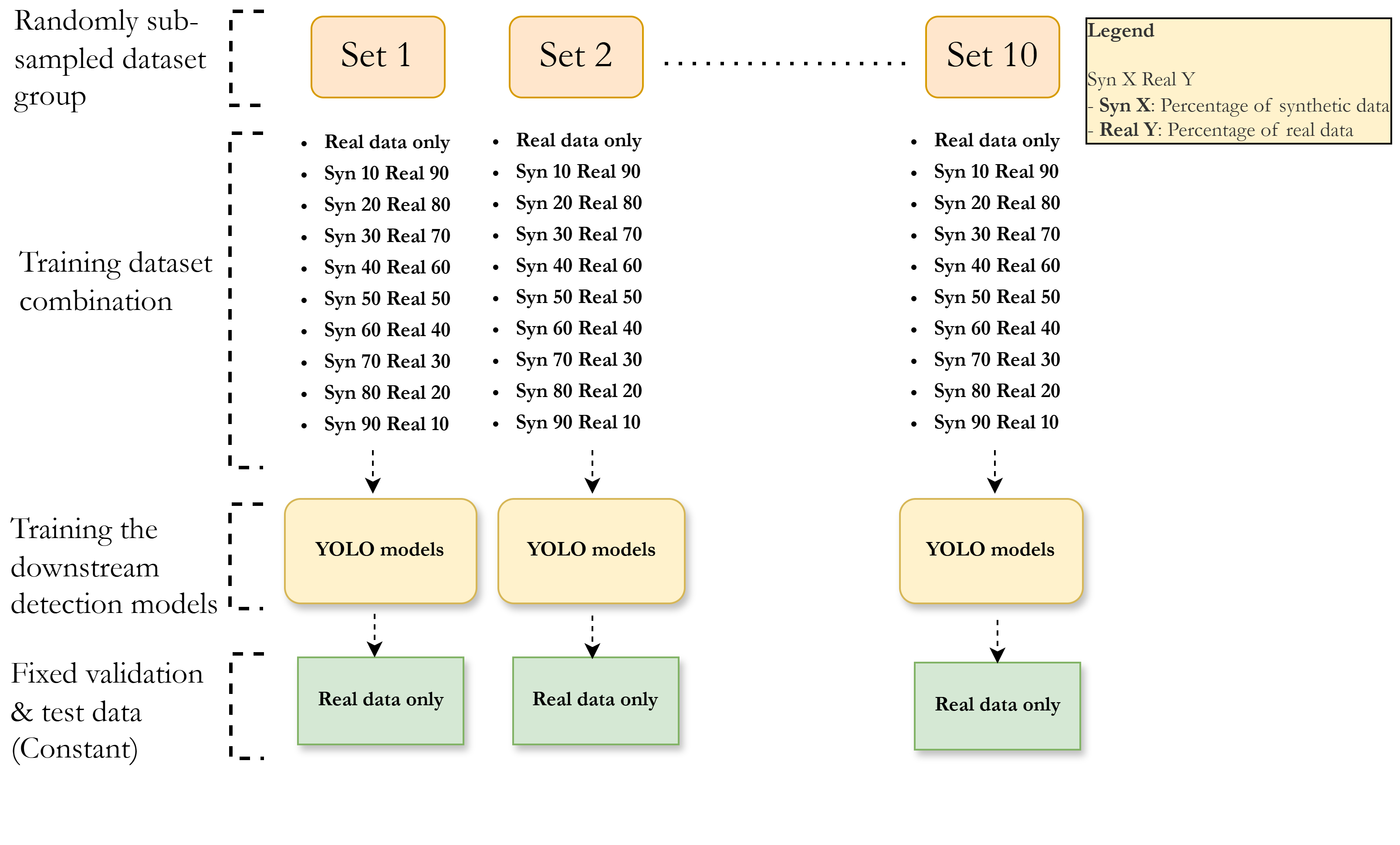}
    \caption{Overview of the targeted random sub-sampling strategy used in this study. Ten independent dataset subsets were generated, each with defined proportions of synthetic (Syn) and real (Real) images. YOLO models were trained on the specific dataset combinations within each subset and evaluated against a fixed validation and test set comprising only real-world data, providing a robust and consistent framework for comparing model performance across varying synthetic-to-real training ratios}
    \label{fig:sub}
\end{figure}

To minimize bias in evaluating the impact of synthetic images on the performance of weed detection models, we implemented a targeted random sub-sampling approach to validate the effects of synthetic training data fractions on the overall performance of the downstream model (see Fig.~\ref{fig:sub}). Specifically, we generated ten independent random sets of the training data, each containing predefined mixtures of real and synthetic images in varying proportions (e.g., 10\% synthetic, 90\% real; 20\% synthetic, 80\% real, etc.). The validation and test datasets were kept constant, consisting solely of real data. We trained state-of-the-art YOLO models (YOLOv8n, YOLOv8s, YOLOv9t, YOLOv9s, YOLOv10n, and YOLOv10s) on the corresponding training sets. After training, the models were evaluated on a fixed holdout test set composed only of real data to assess their generalization performance on unseen real-world data. The primary evaluation metric was the mAP50 and mAP50\text{-}95.

We begin our statistical analysis by conducting the Shapiro-Wilk test to assess the normality of performance metrics (mAP50 and mAP50\text{-}95) across each dataset combination for each model approach over 10 random sets, setting a significance level of $\alpha = 0.05$ for all analyses. If the data are normally distributed, we proceed with a one-way ANOVA at $\alpha = 0.05$ to examine statistically significant differences in model performance among dataset combinations. Upon detecting significant differences with ANOVA, Tukey’s Honest Significant Difference (HSD) test is applied for post-hoc pairwise comparisons. Tukey’s HSD controls the family-wise error rate, maintaining $\alpha = 0.05$ across all comparisons. For data not meeting normality assumptions, we employ the Kruskal-Wallis test at $\alpha = 0.05$ to detect significant performance differences across dataset combinations. In cases where the Kruskal-Wallis test is significant, we apply Dunn’s test with a Bonferroni correction to control for multiple comparisons during post-hoc analysis at $\alpha = 0.05$. To facilitate interpretation, we assign letter labels in superscript to dataset combinations. Combinations without significant performance differences share the same letter, while those with significant differences are assigned distinct letters, visually summarizing post-hoc comparisons.

\section{Result}
\label{sec:result} 
\begin{figure}[!htbp]
    \centering
    \begin{minipage}{0.49\columnwidth}
        \centering
        \includegraphics[width=\textwidth]{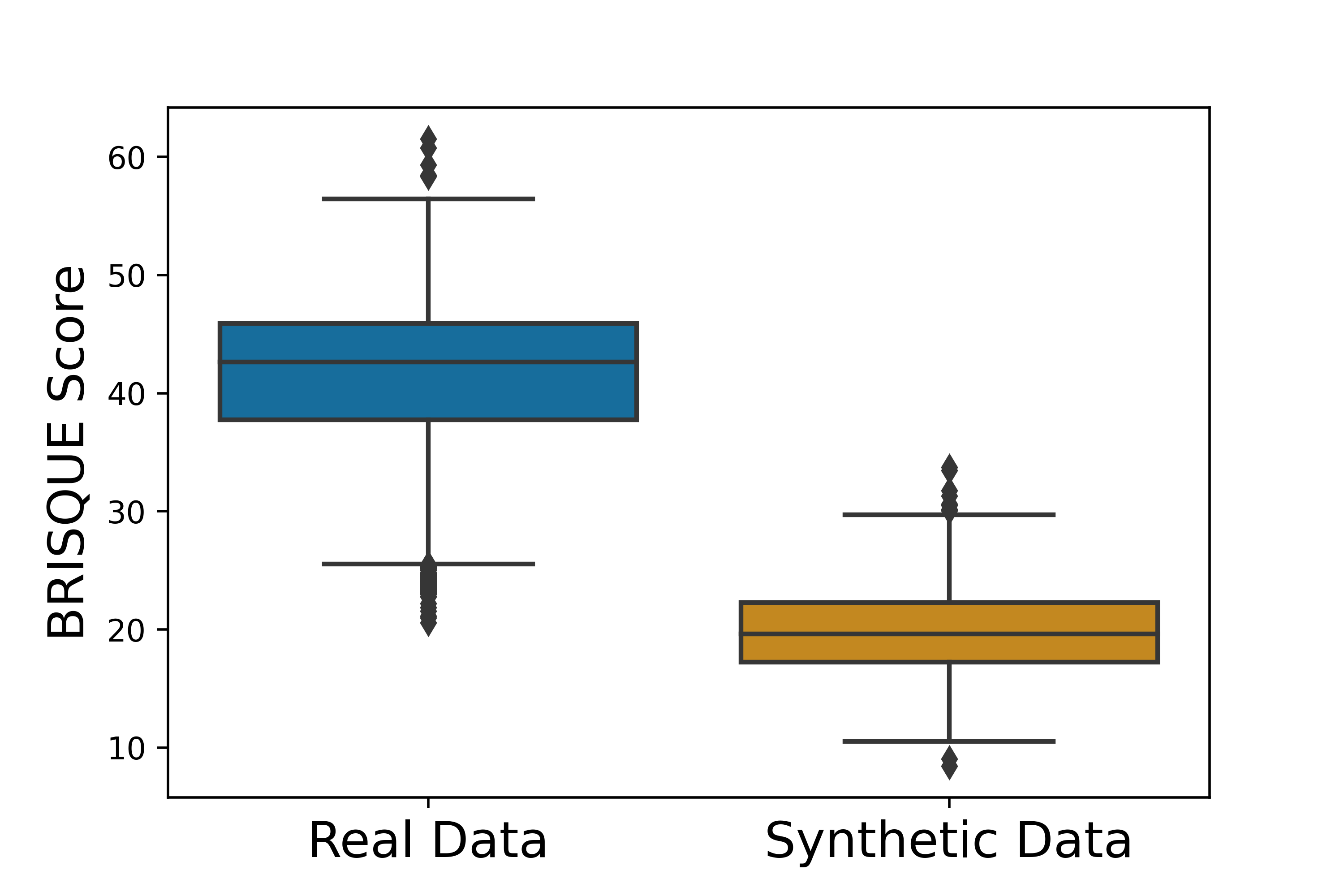} 
        \subcaption{BRISQUE}
        \label{fig:brisque}
    \end{minipage}\hfill
    \begin{minipage}{0.49\columnwidth}
        \centering
        \includegraphics[width=\textwidth]{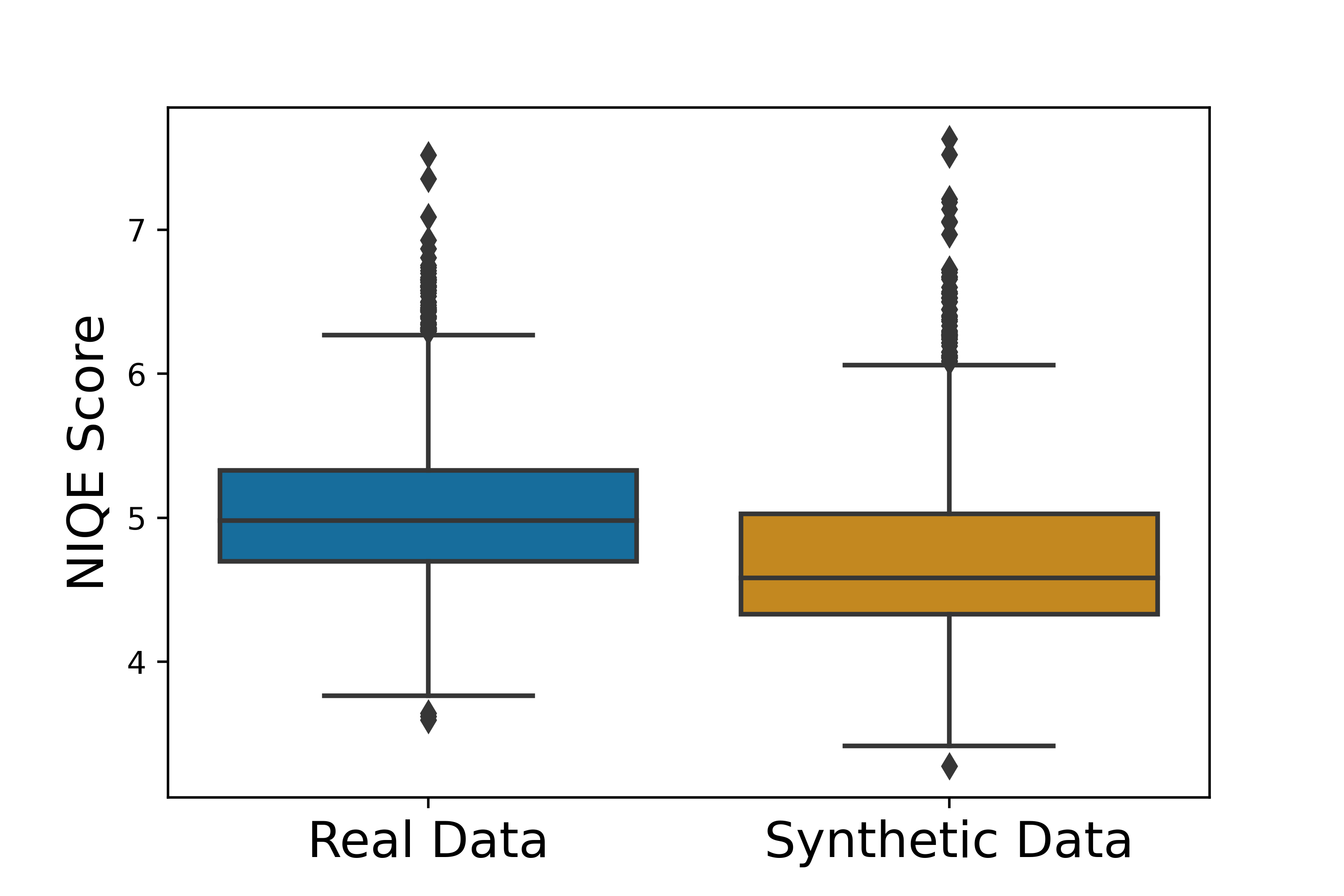} 
        \subcaption{NIQE}
        \label{fig:Niqe}
    \end{minipage}
    \\[1em]
    \begin{minipage}{0.49\columnwidth}
        \centering
        \includegraphics[width=\textwidth]{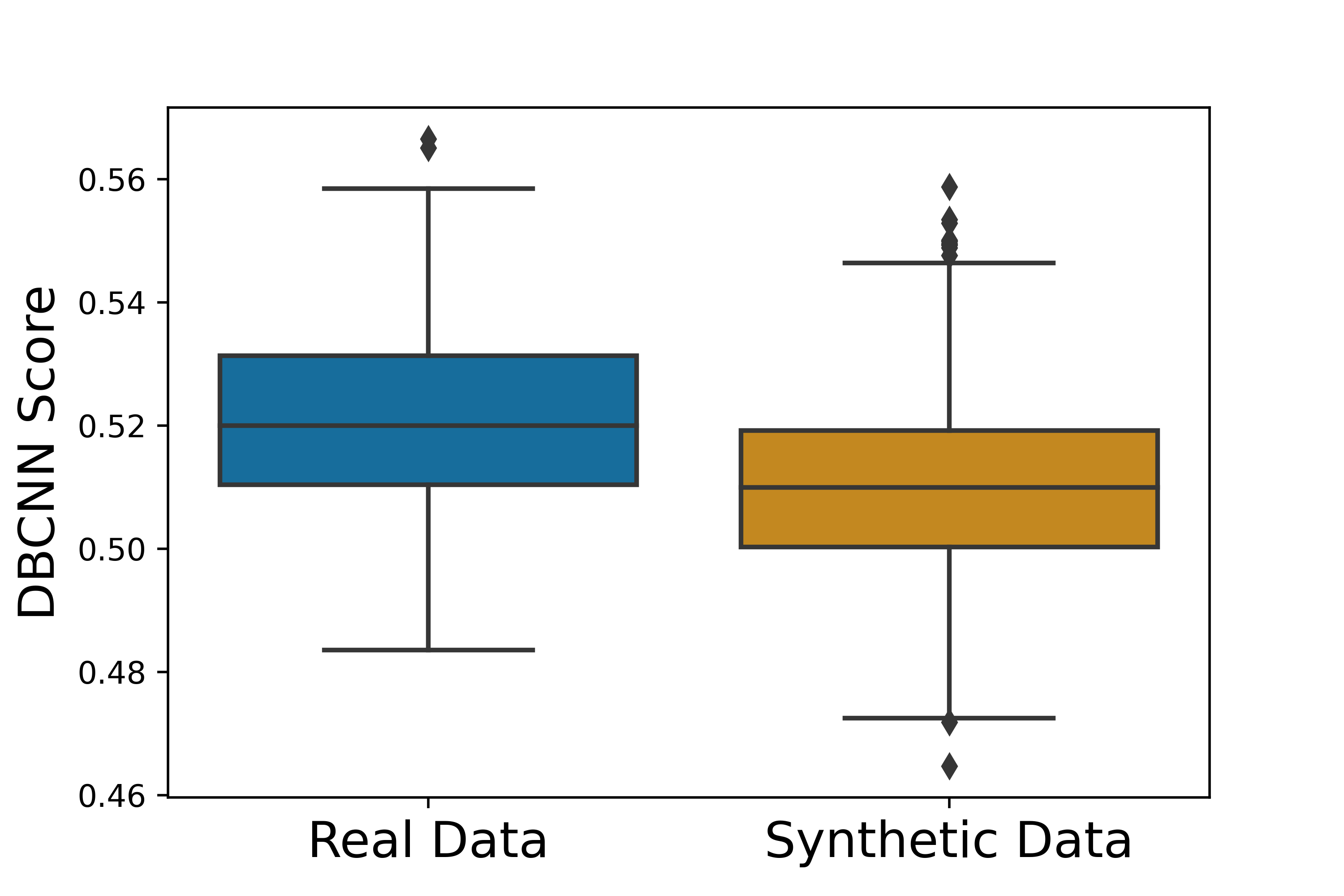} 
        \subcaption{DBCNN}
        \label{fig:dbcnn}
    \end{minipage}\hfill
    \begin{minipage}{0.49\columnwidth}
        \centering
        \includegraphics[width=\textwidth]{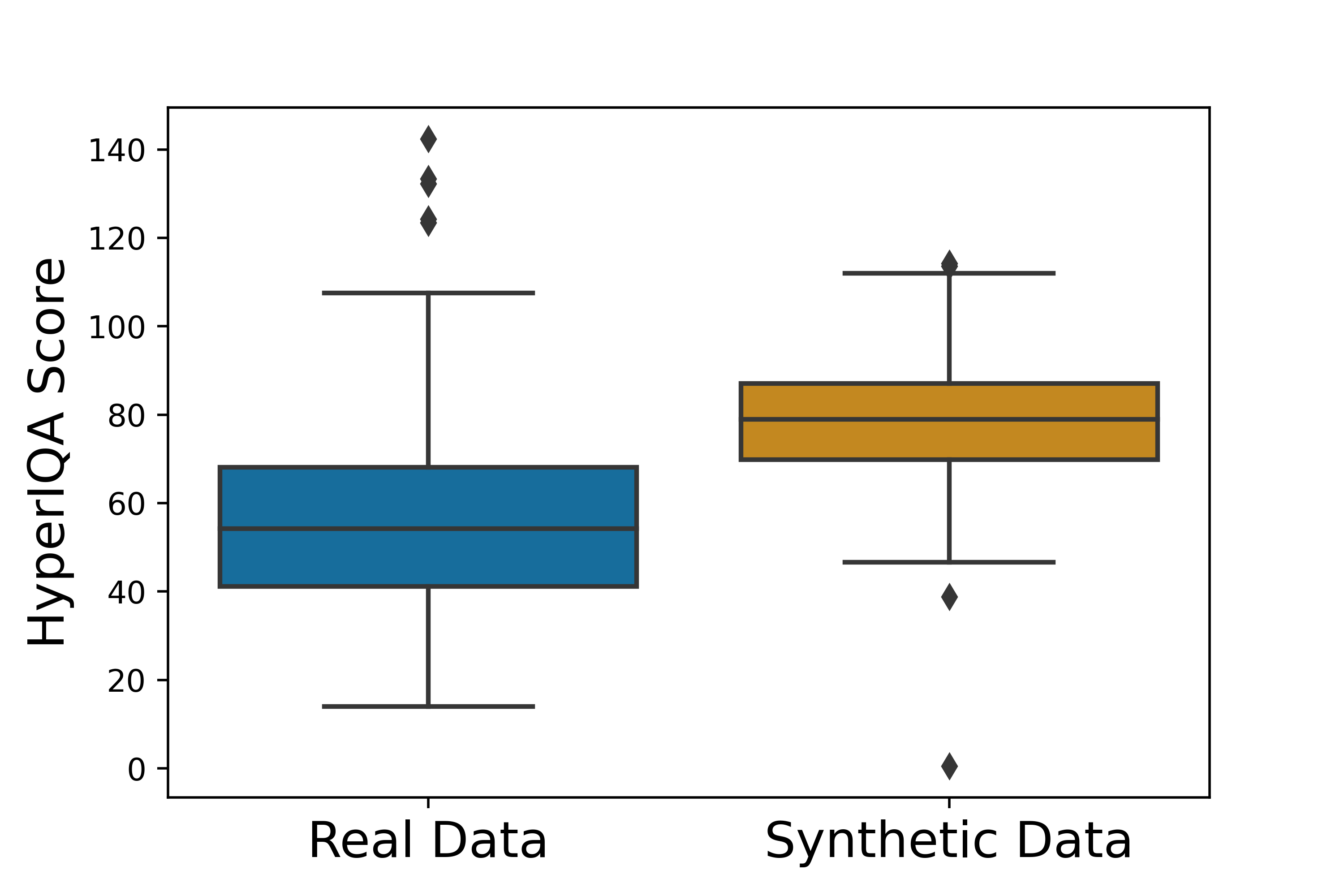} 
        \subcaption{HyperIQA}
        \label{fig:hyper}
    \end{minipage}
    \caption{Boxplots of four NR-IQA metrics (BRISQUE, NIQE, DBCNN, HyperIQA), comparing the image quality between real and synthetic images. For (a) BRISQUE and (b) NIQE, a lower score indicates better image quality, while for (c) DBCNN and (d) HyperIQA, a higher score signifies better image quality. }
    \label{fig:NRIQAs}
\end{figure}

\begin{figure}[!htbp]
    \centering
    \includegraphics[width=01.0\columnwidth,keepaspectratio]{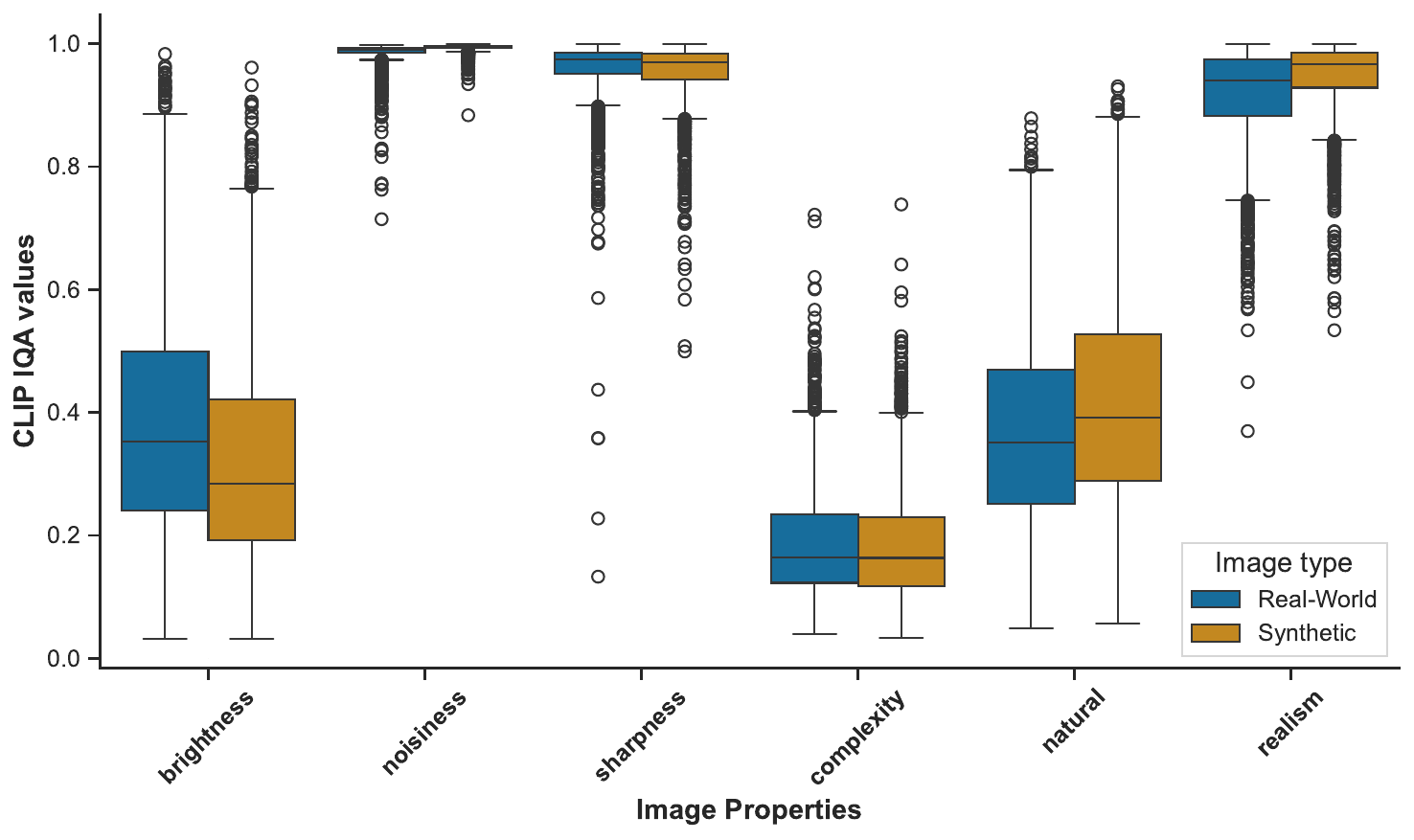}
    \caption{Box plot comparing various image properties (\textit{brightness, noisiness, sharpness, complexity, naturalness}, and \textit{realism}) between real and synthetic images generated with our proposed pipeline}
    \label{fig:clip_iqa}
\end{figure}

To quantitatively evaluate the quality of our synthetic images in comparison to real-world images, we employed the BRISQUE, NIQE, DBCNN, HyperIQA, and CLIP IQA metrics (refer to Fig.~\ref{fig:NRIQAs} and \ref{fig:clip_iqa}). All metrics demonstrated a statistically significant difference ($p < 0.05$) between synthetic and real-world images. For the BRISQUE metric (see Fig.~\ref{fig:brisque}), where lower values indicate better quality, synthetic images ($19.89\pm3.63$) had significantly higher quality than real-world images ($40.88\pm7.30$). Similarly, the NIQE metric (see Fig.\ref{fig:Niqe}), which also favors lower values for higher quality, showed that synthetic images ($4.72\pm0.56$) had slightly better image quality in terms of natural scene statistics compared to real-world images ($5.04\pm0.51$). Conversely, for the DBCNN metric (see Fig.~\ref{fig:dbcnn}), where higher values indicate better quality, real-world images ($0.52\pm0.014$) had slightly better quality than synthetic images ($0.50\pm0.013$). In contrast, the Hyper IQA metric(see Fig.~\ref{fig:hyper}), which also favors higher values for better quality, demonstrated that synthetic images ($78.56\pm11.89$) performed significantly better than real-world images ($56.03\pm18.65$). 

According to CLIP-IQA, the spatial attributes of synthetic images, including \textit{brightness} ($0.323 \pm 0.169$), \textit{noisiness} ($0.993 \pm 0.005$), and \textit{sharpness} ($0.954 \pm 0.048$), exhibit a similar pattern when compared with the real image counterparts, which score as follows: \textit{brightness} ($0.385 \pm 0.186$), \textit{noisiness} ($0.985 \pm 0.0184$), and \textit{sharpness} ($0.958 \pm 0.052$) (cf. Figure \ref{fig:clip_iqa}). Of particular interest are the \textit{natural} and \textit{realism} scores, where interestingly it can be observed, that synthetic images outscore real images. The mean \textit{natural} and \textit{realism} scores of synthetic images evaluate ($0.417 \pm 0.174$) and ($0.946 \pm 0.058$) respectively, surpassing their real-world counterparts, which score ($0.369 \pm 0.153$) and ($0.916 \pm 0.078$) respectively. This would indicate a higher fidelity of the synthetic images.
However, the interpretation of this observation is not straightforward. Since these metrics are no-reference metrics, and thus are calculated independently and not in pairs of a synthetic image with a reference real image, a conclusion that the synthetic images in our scenario have a higher fidelity than real images in general cannot be drawn.
Additionally, the \textit{complexity} score of synthetic images ($0.187 \pm 0.0916$) aligns closely with that of real images ($0.182 \pm 0.0884$) (cf. Figure \ref{fig:clip_iqa}). Following the statistical test, the results indicate significant differences in the \textit{brightness}, \textit{noisiness}, \textit{sharpness}, \textit{natural}, and \textit{realism} values with $p$-values $< 0.05$; only the \textit{complexity} values showed no significant differences ($p > 0.05$).



To evaluate the effectiveness of synthetic images in downstream models, we trained several versions of YOLO nano (YOLOv8n, YOLOv9t, YOLOv10n) and YOLO small (YOLOv8s, YOLOv9s, YOLOv10s) across a range of datasets containing both real and mixed (real and synthetic) data. The trained models were tested using real-world data to evaluate their performance in real-world environments and assess their \textit{data efficiency}.

The \textit{mAP50} scores for the various YOLO models (see Tab.~\ref{tab:yolo_mAP50}) trained on real and mixed datasets reveal clear trends, highlighting the impact of synthetic data integration on model performance. Introducing a small proportion of synthetic data generally improved model performance. For most models, the Syn10 Real90  dataset produced the highest mAP50 scores, suggesting that adding 10\% synthetic data can enhance performance. For example, YOLOv8n achieved a mean mAP50 score of \(0.883 \pm 0.007\) with 10\% synthetic images, showing no significant differences from its performance in real data only \((p > 0.05)\). This indicates that introducing 10\% synthetic data does not negatively impact the model's ability to detect real-world data. However, as the proportion of synthetic data increased beyond 10\%, YOLOv8n began to show declines in performance. Its performance dropped significantly on the Syn40 Real60 dataset, with a mean mAP50 score of \(0.854 \pm 0.008\).

Models, such as YOLOv9s, effectively integrated synthetic data at moderate levels without a significant decline in performance. For example, YOLOv9s achieved a mean mAP50 score of \(0.887 \pm 0.009\) with 30\% synthetic data, with no significant difference compared to its performance in real data only \((p > 0.05)\). This demonstrates the model's ability to incorporate synthetic data effectively up to a certain threshold without experiencing a significant accuracy loss.

While most models showed performance declines at higher synthetic data levels, the YOLOv10 model family exhibited high resilience to larger proportions of synthetic data. YOLOv10n maintained a mean mAP50 score of \(0.732 \pm 0.027\) with 90\% synthetic data, showing no significant difference compared to real data only \((p > 0.05)\). Similarly, YOLOv10s maintained strong baseline performance even with up to 80\% synthetic data, achieving a mean mAP50 score of \(0.790 \pm 0.019\) without significant differences from real data alone. Furthermore, YOLOv10s outperformed other data set combinations in the Syn10 Real90 dataset, achieving a mean mAP50 score of \(0.859 \pm 0.013\), which was statistically superior to performance only with real data \((p < 0.05\)).

In contrast, some models experienced significant performance drops as the proportion of synthetic data increased. For instance, the performance of YOLOv8s dropped significantly to \(0.877 \pm 0.008\) with just 20\% synthetic data compared to its real data only\(0.892 \pm 0.000\) counterpart \((p < 0.05)\). A similar pattern was observed with YOLOv9t, where its performance dropped significantly \((p < 0.05)\) from \(0.902 \pm 0.000\) with real data to \(0.889 \pm 0.008\) with just 10\% synthetic data.

\begin{table}[htbp]
    \centering
    \caption{Comparison of mAP50 scores (mean $\pm$ 1SD) across various YOLO models—YOLO nano (YOLOv8n, YOLOv9t, YOLOv10n) and YOLO small (YOLOv8s, YOLOv9s, YOLOv10s)—on real and mixed datasets (combinations of SynX and Real Y), along with statistical significance groupings (groups indicated as superscript letters). Datasets with the same letter are not significantly different, while those with different letters show statistically significant differences. The highest mAP50 score in each model column is highlighted in bold.}
\resizebox{\textwidth}{!}{%
 \begin{tabular}{lllllllll}
\hline
\textbf{Dataset} & \makecell{\textbf{YOLOv8n} \\ (mean $\pm$ 1SD)} & \makecell{\textbf{YOLOv8s} \\ (mean $\pm$ 1SD)} & \makecell{\textbf{YOLOv9t} \\ (mean $\pm$ 1SD)} & \makecell{\textbf{YOLOv9s} \\ (mean $\pm$ 1SD)} & \makecell{\textbf{YOLOv10n} \\ (mean $\pm$ 1SD)} & \makecell{\textbf{YOLOv10s} \\ (mean $\pm$ 1SD)} \\
\hline
\addlinespace[2pt]
Real data only & 0.876 ± 0.000\textsuperscript{A} & \textbf{0.892} ± 0.000\textsuperscript{A} & \textbf{0.902} ± 0.000\textsuperscript{A} & 0.887 ± 0.000\textsuperscript{A} & 0.787 ± 0.000\textsuperscript{A} & 0.799 ± 0.000\textsuperscript{A} \\
Syn10 Real90  & \textbf{0.883} ± 0.007\textsuperscript{A} & 0.886 ± 0.008\textsuperscript{B} & 0.889 ± 0.008\textsuperscript{B} & \textbf{0.899} ± 0.009\textsuperscript{A} & \textbf{0.823} ± 0.018\textsuperscript{A} & \textbf{0.859} ± 0.013\textsuperscript{B} \\
Syn20 Real80  & 0.870 ± 0.006\textsuperscript{B} & 0.877 ± 0.008\textsuperscript{B} & 0.878 ± 0.010\textsuperscript{B} & 0.890 ± 0.010\textsuperscript{A} & 0.818 ± 0.015\textsuperscript{A} & 0.845 ± 0.018\textsuperscript{C} \\
Syn30 Real70  & 0.869 ± 0.014\textsuperscript{B} & 0.879 ± 0.010\textsuperscript{B} & 0.867 ± 0.012\textsuperscript{B} & 0.887 ± 0.009\textsuperscript{A} & 0.812 ± 0.018\textsuperscript{A} & 0.836 ± 0.015\textsuperscript{D} \\
Syn40 Real60  & 0.854 ± 0.008\textsuperscript{B} & 0.862 ± 0.011\textsuperscript{B} & 0.859 ± 0.010\textsuperscript{B} & 0.877 ± 0.009\textsuperscript{B} & 0.805 ± 0.021\textsuperscript{A} & 0.829 ± 0.014\textsuperscript{E} \\
Syn50 Real50  & 0.850 ± 0.010\textsuperscript{B} & 0.860 ± 0.014\textsuperscript{B} & 0.856 ± 0.011\textsuperscript{B} & 0.868 ± 0.006\textsuperscript{B} & 0.780 ± 0.030\textsuperscript{A} & 0.821 ± 0.022\textsuperscript{A} \\
Syn60 Real40  & 0.838 ± 0.010\textsuperscript{B} & 0.852 ± 0.009\textsuperscript{C} & 0.847 ± 0.012\textsuperscript{C} & 0.855 ± 0.014\textsuperscript{B} & 0.792 ± 0.025\textsuperscript{A} & 0.809 ± 0.022\textsuperscript{A} \\
Syn70 Real30  & 0.820 ± 0.018\textsuperscript{B} & 0.836 ± 0.012\textsuperscript{C} & 0.838 ± 0.011\textsuperscript{C} & 0.856 ± 0.011\textsuperscript{B} & 0.781 ± 0.025\textsuperscript{A} & 0.800 ± 0.029\textsuperscript{A} \\
Syn80 Real20  & 0.814 ± 0.011\textsuperscript{B} & 0.822 ± 0.016\textsuperscript{C} & 0.829 ± 0.008\textsuperscript{C} & 0.843 ± 0.008\textsuperscript{B} & 0.764 ± 0.019\textsuperscript{A} & 0.790 ± 0.019\textsuperscript{A} \\
Syn90 Real10  & 0.780 ± 0.011\textsuperscript{B} & 0.794 ± 0.013\textsuperscript{C} & 0.799 ± 0.010\textsuperscript{C} & 0.811 ± 0.006\textsuperscript{B} & 0.732 ± 0.027\textsuperscript{A} & 0.760 ± 0.017\textsuperscript{F} \\
\hline
\end{tabular}
}
    \label{tab:yolo_mAP50}
\end{table}

Turning to the \textit{mAP50\text{-}95} metrics for the different YOLO models (see Tab.~\ref{tab:yolo_mAP50_95}), a pattern similar to the mAP50 scores emerges, with the inclusion of a small proportion of synthetic data that generally does not hamper and in some cases improve model performance. Most YOLO models achieved their highest average mAP50\text{-}95 scores on datasets with 10\% synthetic images (Syn10 Real90). For example, YOLOv8n achieved an observed mean mAP50\text{-}95 score of \(0.713 \pm 0.006\) on the Syn10 Real90 dataset, which was marginally higher than its performance on real data alone \((0.687 \pm 0.000)\); however, this difference was not statistically significant \((p > 0.05)\). Similarly, YOLOv8s exhibited a positive trend, with a mean mAP50\text{-}95 increasing from \(0.691 \pm 0.000\) on real data to \(0.730 \pm 0.007\) with 10\% synthetic data, though this increase was also non-significant \((p > 0.05)\). YOLOv9t followed a similar pattern with 10\% of synthetic data, increasing from only real data \(0.697 \pm 0.000\) to \(0.714 \pm 0.007\), but this difference was also not statistically significant \((p > 0.05)\). 

In contrast, YOLOv9s demonstrated a statistically significant performance improvement with the addition of 10\% synthetic data, achieving a mean mAP50\text{-}95 increase from \(0.684 \pm 0.000\) on real data alone to \(0.735 \pm 0.007\) \((p < 0.05)\). Similarly, YOLOv10s exhibited a statistically significant improvement in mAP50\text{-}95, increasing from \(0.601 \pm 0.000\) on real images to \(0.699 \pm 0.018\) with 10\% synthetic data \((p < 0.05)\). 

Beyond 10\% synthetic data, most YOLO models maintained performance comparable to their real-data counterparts, even with moderate increases in synthetic data. For example, YOLOv8n performed similarly with the 30\% of synthetic data with no statistically significant difference compared to the trained only with real data \((p > 0.05)\). Similarly, YOLOv8s and YOLOv9t maintained strong performance with 50\% and 20\% synthetic data, respectively, with no significant difference from those trained with only real data \((p > 0.05)\).

YOLOv9s exhibited impressive resilience to increasing proportions of synthetic data, showing no statistically significant performance decline even when trained with up to 90\% synthetic data compared to training exclusively on real data \((p > 0.05)\). YOLOv9s achieved its highest mean mAP50\text{-}95 score of \(0.735 \pm 0.007\) with 10\% synthetic data and maintained similar performance up to 50\% synthetic data, with no statistically significant differences observed \((p > 0.05)\). Similarly, both YOLOv10n and YOLOv10 demonstrated strong resilience in synthetic data, maintaining performance consistent with baseline levels when trained with up to 90\% synthetic data, without statistically significant differences from baseline with only real data \((p > 0.05)\). Besides, when trained with 20\% synthetic data, YOLOv10n significantly outperformed the baseline performance, achieving a mean mAP50\text{-}95 score of \(0.640 \pm 0.018\), compared to \(0.590 \pm 0.000\) for the model trained solely on real data \((p < 0.05)\). Moreover, it maintained consistent performance up to 70\% synthetic data, with no statistically significant differences observed the highest score with only 20\% of synthetic data \((p > 0.05)\). Similarly, YOLOv10s demonstrated strong adaptability to synthetic data, surpassing their performance with only real data. Specifically, YOLOv10s achieved a mean mAP50\text{-}95 score of \(0.699 \pm 0.018\) with 10\% synthetic data, compared to \(0.601 \pm 0.000\) with real data alone \((p < 0.05)\). Even in 50\% synthetic data, YOLOv10s maintained a strong mAP50\text{-}95 score of \(0.665 \pm 0.022\), with no significant difference from its peak performance at 10\% synthetic data \((p > 0.05)\).


\begin{table}[htbp]
    \centering
    \caption{Comparison of mAP50\text{-}95 scores (mean $\pm$ 1SD) across various YOLO models—YOLO nano (YOLOv8n, YOLOv9t, YOLOv10n) and YOLO small (YOLOv8s, YOLOv9s, YOLOv10s)—on real and mixed datasets (combinations of SynX and Real Y), along with statistical significance groupings (groups indicated as superscript letters). Datasets with the same letter are not significantly different, while those with different letters show statistically significant differences. The highest mAP50\text{-}95 score in each model column is highlighted in bold.}
\resizebox{\textwidth}{!}{%
   \begin{tabular}{lllllllll}
\hline
\textbf{Dataset} & \makecell{\textbf{YOLOv8n} \\ (mean $\pm$ 1SD)} & \makecell{\textbf{YOLOv8s} \\ (mean $\pm$ 1SD)} & \makecell{\textbf{YOLOv9t} \\ (mean $\pm$ 1SD)} & \makecell{\textbf{YOLOv9s} \\ (mean $\pm$ 1SD)} & \makecell{\textbf{YOLOv10n} \\ (mean $\pm$ 1SD)} & \makecell{\textbf{YOLOv10s} \\ (mean $\pm$ 1SD)} \\
\hline
\addlinespace[2pt]
       Real data only &          0.687 ± 0.000\textsuperscript{A} &          0.691 ± 0.000\textsuperscript{A} &          0.697 ± 0.000\textsuperscript{A} &          0.684 ± 0.000\textsuperscript{A} &          0.590 ± 0.000\textsuperscript{A} &          0.601 ± 0.000\textsuperscript{A} \\
Syn10 Real90  & \textbf{0.713} ± 0.006\textsuperscript{A} & \textbf{0.730} ± 0.007\textsuperscript{A} & \textbf{0.714} ± 0.007\textsuperscript{A} & \textbf{0.735} ± 0.007\textsuperscript{B} &          0.635 ± 0.014\textsuperscript{B} & \textbf{0.699} ± 0.018\textsuperscript{B} \\
Syn20 Real80  &          0.702 ± 0.007\textsuperscript{A} &          0.719 ± 0.007\textsuperscript{A} &          0.699 ± 0.007\textsuperscript{A} &          0.727 ± 0.007\textsuperscript{B} & \textbf{0.640} ± 0.018\textsuperscript{B} &          0.689 ± 0.016\textsuperscript{B} \\
Syn30 Real70  &          0.690 ± 0.008\textsuperscript{A} &          0.714 ± 0.011\textsuperscript{A} &          0.689 ± 0.007\textsuperscript{B} &          0.719 ± 0.012\textsuperscript{B} &          0.629 ± 0.013\textsuperscript{B} &          0.682 ± 0.017\textsuperscript{B} \\
Syn40 Real60  &          0.681 ± 0.006\textsuperscript{B} &          0.701 ± 0.008\textsuperscript{A} &          0.684 ± 0.012\textsuperscript{B} &          0.710 ± 0.008\textsuperscript{B} &          0.629 ± 0.017\textsuperscript{B} &          0.673 ± 0.014\textsuperscript{B} \\
Syn50 Real50  &          0.668 ± 0.007\textsuperscript{B} &          0.690 ± 0.009\textsuperscript{A} &          0.673 ± 0.006\textsuperscript{B} &          0.702 ± 0.012\textsuperscript{B} &          0.608 ± 0.020\textsuperscript{B} &          0.665 ± 0.022\textsuperscript{B} \\
Syn60 Real40  &          0.656 ± 0.010\textsuperscript{B} &          0.681 ± 0.007\textsuperscript{B} &          0.664 ± 0.008\textsuperscript{B} &          0.691 ± 0.010\textsuperscript{A} &          0.616 ± 0.031\textsuperscript{B} &          0.646 ± 0.024\textsuperscript{A} \\
Syn70 Real30  &          0.644 ± 0.010\textsuperscript{B} &          0.664 ± 0.012\textsuperscript{B} &          0.658 ± 0.007\textsuperscript{B} &          0.683 ± 0.007\textsuperscript{A} &          0.602 ± 0.019\textsuperscript{B} &          0.638 ± 0.027\textsuperscript{A} \\
Syn80 Real20  &          0.627 ± 0.008\textsuperscript{B} &          0.646 ± 0.009\textsuperscript{B} &          0.639 ± 0.004\textsuperscript{B} &          0.671 ± 0.007\textsuperscript{A} &          0.582 ± 0.023\textsuperscript{A} &          0.621 ± 0.019\textsuperscript{A} \\
Syn90 Real10  &          0.596 ± 0.008\textsuperscript{B} &          0.610 ± 0.006\textsuperscript{B} &          0.606 ± 0.010\textsuperscript{B} &          0.633 ± 0.007\textsuperscript{A} &          0.547 ± 0.022\textsuperscript{A} &          0.585 ± 0.021\textsuperscript{A} \\
\hline
\end{tabular}
}
    \label{tab:yolo_mAP50_95}
\end{table}

\section{Discussion}
\label{sec: disc}

In the \textit{image-specific} analysis, the results indicate that our synthetic images exhibit a pattern similar to that of real-world images across various NR-IQA metrics, including BRISQUE, NIQE, DBCNN, HyperIQA, and CLIP IQA metrics (cf. Fig.~\ref{fig:NRIQAs} \& ~\ref{fig:clip_iqa}). Notably, metrics such as BRISQUE, NIQE, HyperIQA, and CLIP IQA suggest that synthetic images may have higher quality compared to real-world images. However, the DBCNN metric presents a contrasting result. The DBCNN metric evaluates aspects such as Gaussian blur, white noise, JPEG compression, contrast changes, pink noise, overexposure, and underexposure. Our synthetic images might exhibit one or more of these issues, necessitating further analysis to identify the specific problems. If specific issues are identified, post-processing techniques such as image motion deblurring, denoising, or image super-resolution may be employed to enhance image quality.



The \textit{task-specific} evaluation approach focuses on data-efficient training and utilizes both mAP50 and mAP50\text{-}95 metrics to offer essential insights into the YOLO models' performance in detecting weeds, especially when trained using different mixtures of real and synthetic datasets. A key finding is that the inclusion of a small proportion of synthetic images does not negatively impact on most of the model's performance. In contrast, it improves the performance of some models compared to models trained exclusively on real-world images. This suggests synthetic images introduce beneficial variability that enhances the model's generalization capabilities.

The results from the mAP50 metric (see Tab.~\ref{tab:yolo_mAP50}) show that models such as YOLOv8n, YOLOv9s, YOLOv10n, and YOLOv10s when trained with a small proportion (10\%) of synthetic data, exhibit performance comparable to those trained exclusively on real-world data. Additionally, these models outperform others trained with different combinations of real and synthetic data. Notably, YOLOv9s, YOLOv10n, and YOLOv10s, trained with 30\%, 90\%, and 80\% synthetic data, respectively, demonstrate similar performance. This indicates that the partial replacement of real-world data with synthetic data, depending on the model, does not result in significant accuracy loss.

However, models such as YOLOv8s and YOLOv9t did not show any improvement with the inclusion of synthetic data. A plausible explanation for this could be that the synthetic data is not well suited to these particular model architectures. A study by~\cite{ljungqvist2023object} indicated that certain architectures, such as YOLOv3, exhibit differences in the head / classifier when trained on synthetic versus real data. These differences suggest that certain architectures are better at leveraging the generalized features provided by synthetic data. This may also explain the strong performance of YOLOv10 models, potentially due to their different head or classifier architecture.

Turning to the mAP50\text{-}95 metric (see Tab.~\ref{tab:yolo_mAP50_95}), all models achieved their best performance with the introduction of a small proportion of synthetic data. Unlike mAP50 results, YOLOv8s and YOLOv9t did not exhibit a significant performance drop in  mAP50\text{-}95 metric with increasing moderate amounts of synthetic data. Instead, these models maintained consistent performance with real-world data until the proportion of synthetic data reached 50\% and 20\%, respectively.

Models such as YOLOv9s, YOLOv10n, and YOLOv10s, which demonstrated higher resilience to synthetic data, showed consistently strong performance with up to 90\% synthetic data. Both YOLOv9s and YOLOv10s outperformed models trained exclusively on real-world data when trained with 10\% to 50\% synthetic data. Similarly, YOLOv10n outperformed real-world models when trained with 10\% to 70\% synthetic data. These findings indicate that synthetic images contribute effectively to detection tasks, particularly at higher IoU thresholds (50 to 95), where detecting harder or more obscure objects is crucial. This observation aligns with the findings of Modak et al.~\cite{Modak2024WeedDetectionGenAI}, where the authors demonstrated that the augmentations with synthetic images led to higher mAP50\text{-}95 scores compared to traditional augmentation and real-world data alone. The synthetic images utilized in their research exhibited natural variations, encompassing different weed species, and closely mirrored real-world scenarios. This enhanced diversity probably played a role in enhancing the generalization capabilities of the YOLO models.

 However, while synthetic images enhance performance up to a certain threshold, a higher proportion of synthetic data leads to a decrease in performance. Therefore, real-world images are indispensable and cannot yet be fully replaced by synthetic images generated by the current state of our image generation pipeline. This pattern aligns with the outcomes reported by ~\cite{jelic2021can}, where a comparable reduction in performance was noted when real-world data was substituted with synthetic data.

\begin{figure}[htbp]
    \centering
    \textbf{Human Annotation } \\[0.5em] 
    \begin{subfigure}[b]{0.23\textwidth}
        \centering
        \includegraphics[width=\textwidth]{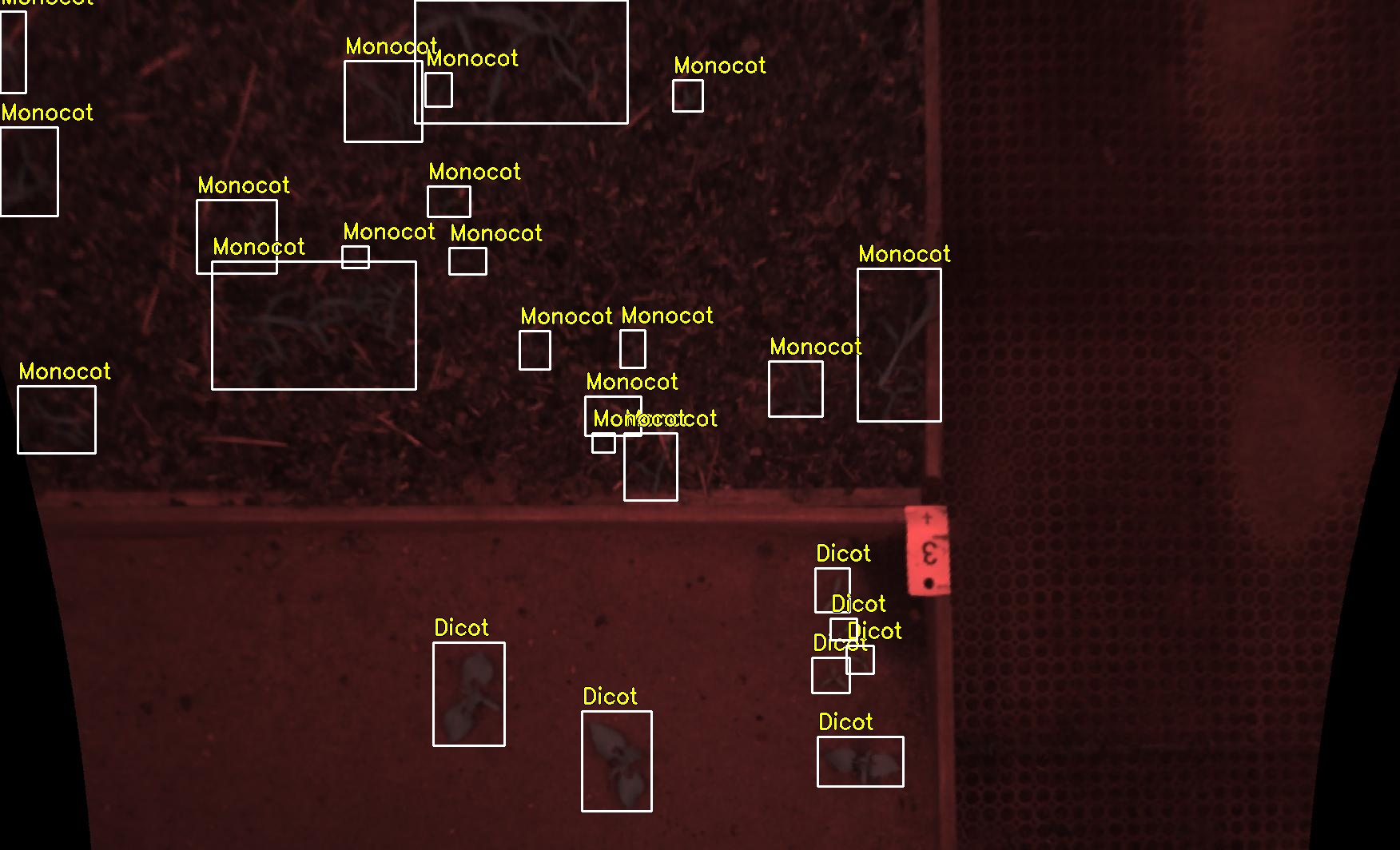}
        \subcaption*{(a)}
    \end{subfigure}
    \hfill
    \begin{subfigure}[b]{0.23\textwidth}
        \centering
        \includegraphics[width=\textwidth]{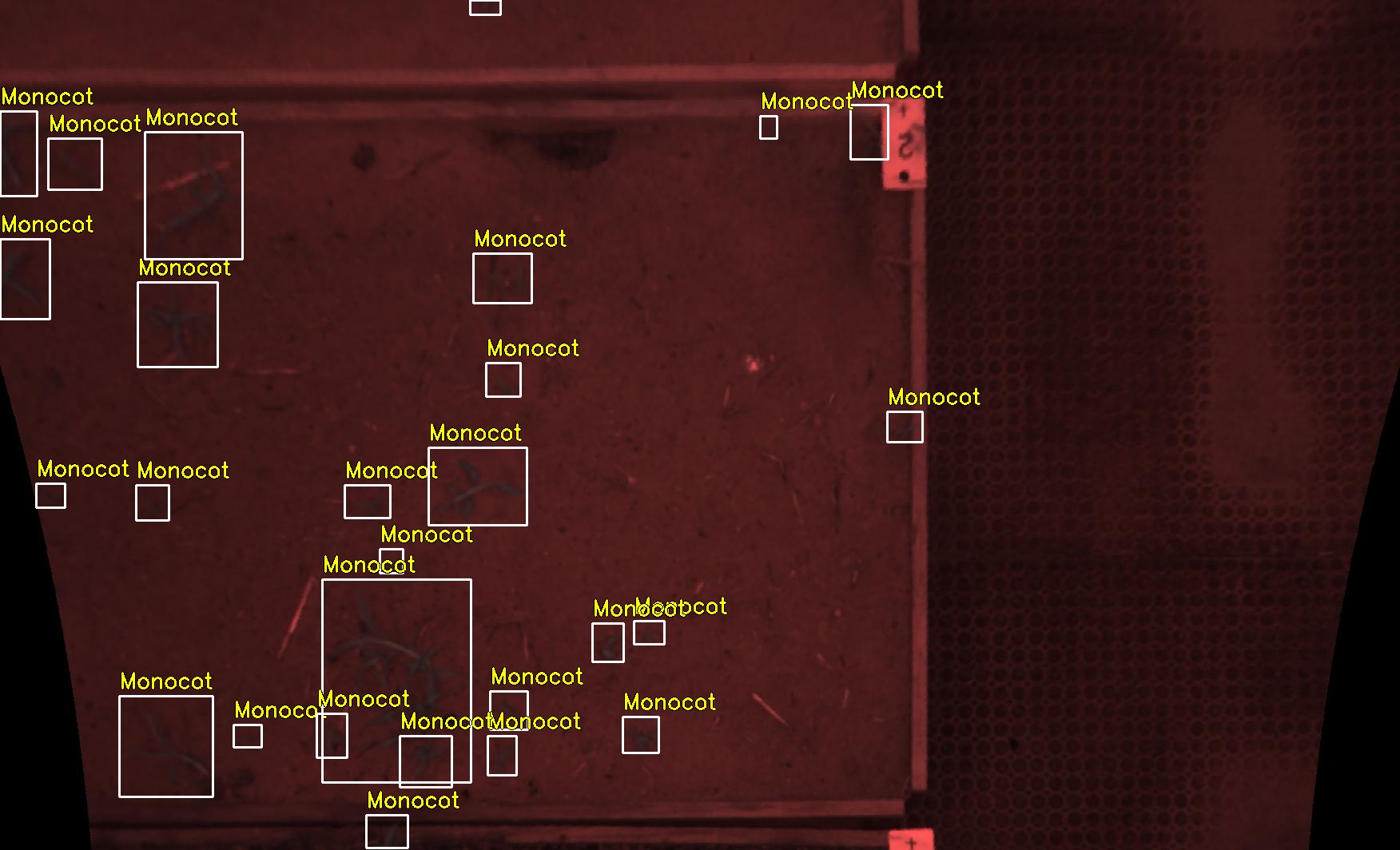}
        \subcaption*{(b)}
    \end{subfigure}
    \hfill
    \begin{subfigure}[b]{0.23\textwidth}
        \centering
        \includegraphics[width=\textwidth]{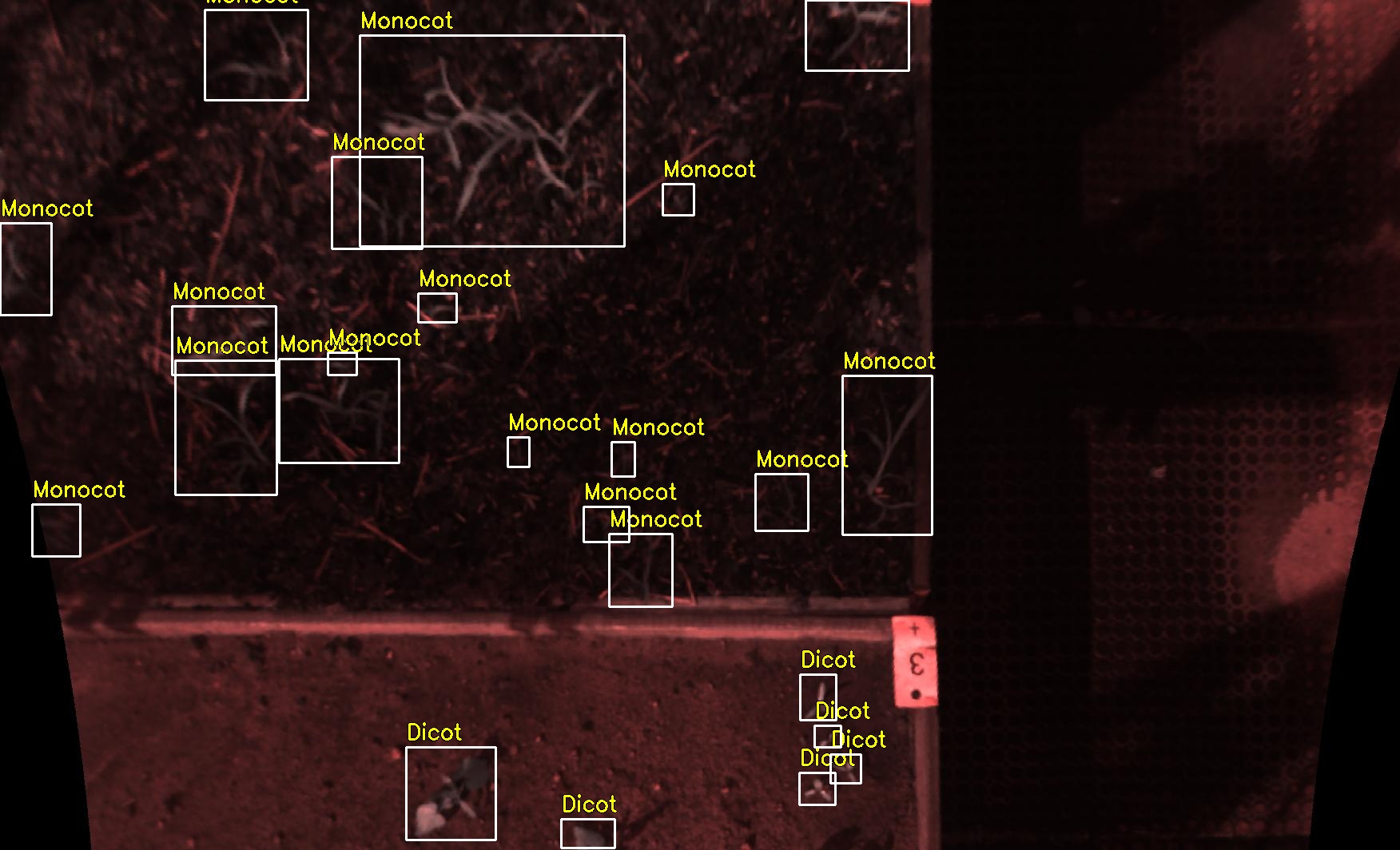}
        \subcaption*{(c)}
    \end{subfigure}
    \hfill
    \begin{subfigure}[b]{0.23\textwidth}
        \centering
        \includegraphics[width=\textwidth]{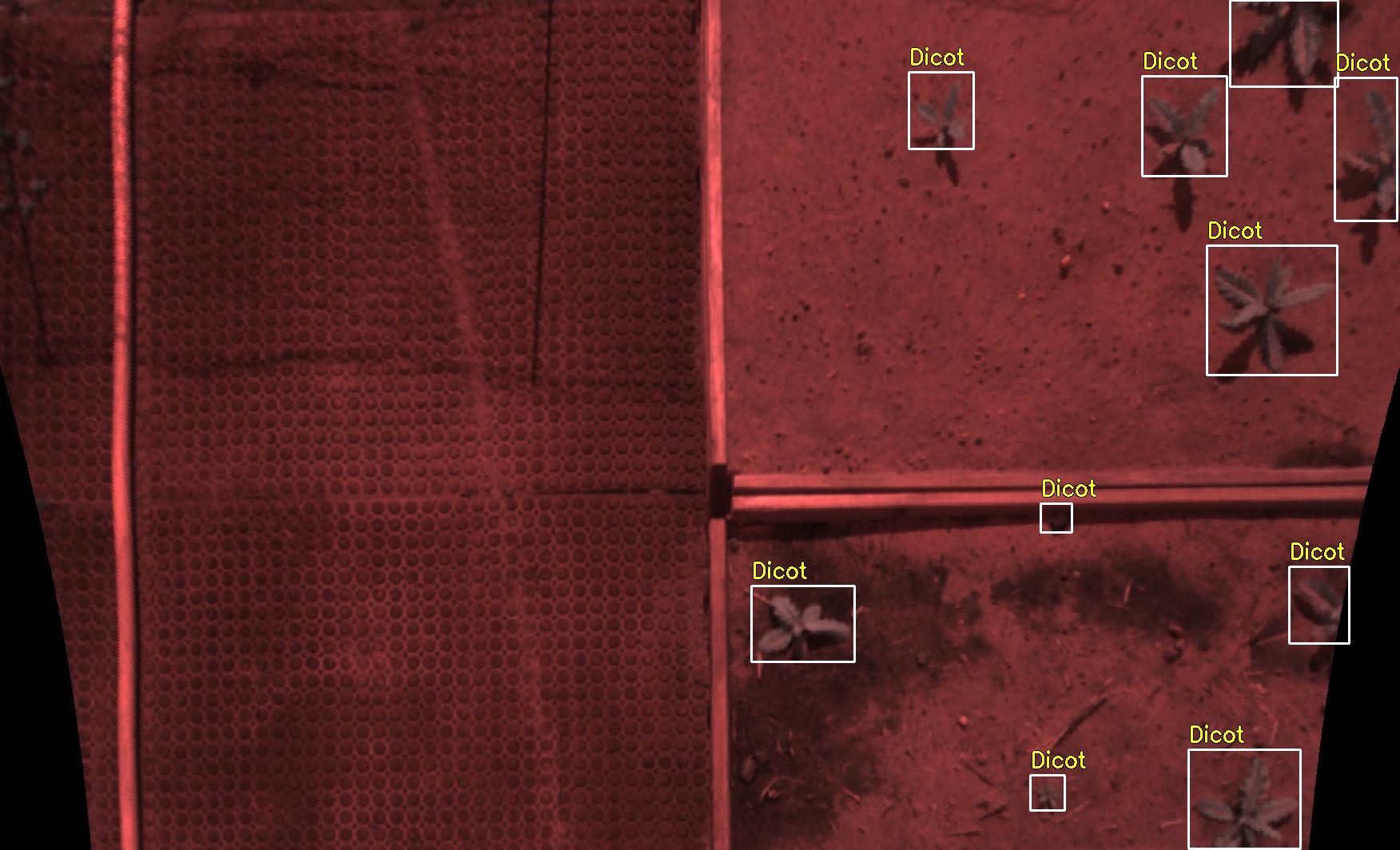}
        \subcaption*{(d)}
    \end{subfigure}

    \vspace{0.8em} 

    \textbf{Model-guided Annotation} \\[0.5em] 
    \begin{subfigure}[b]{0.23\textwidth}
        \centering
        \includegraphics[width=\textwidth]{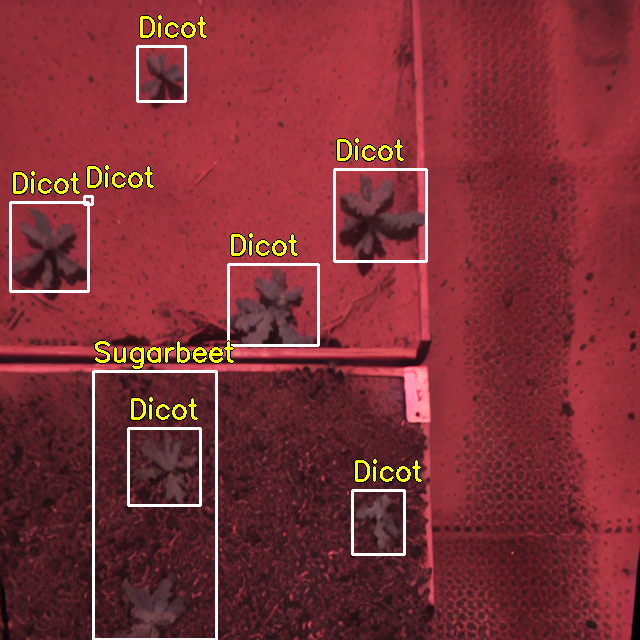}
        \subcaption*{(e)}
    \end{subfigure}
    \hfill
    \begin{subfigure}[b]{0.23\textwidth}
        \centering
        \includegraphics[width=\textwidth]{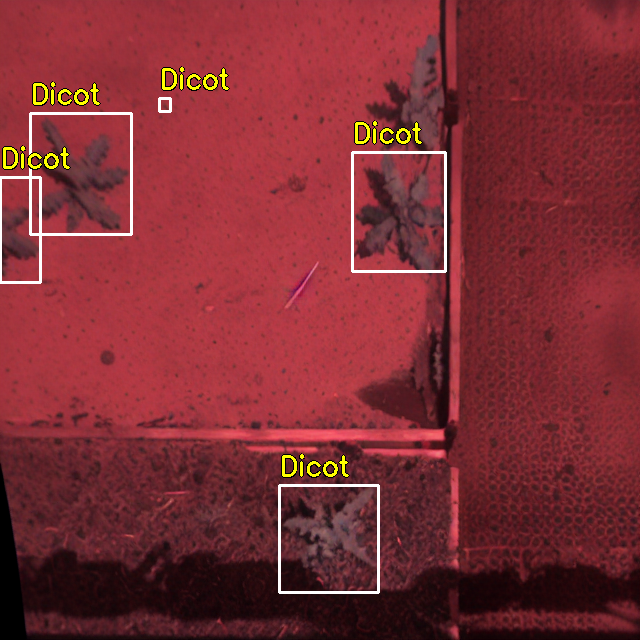}
        \subcaption*{(f)}
    \end{subfigure}
    \hfill
    \begin{subfigure}[b]{0.23\textwidth}
        \centering
        \includegraphics[width=\textwidth]{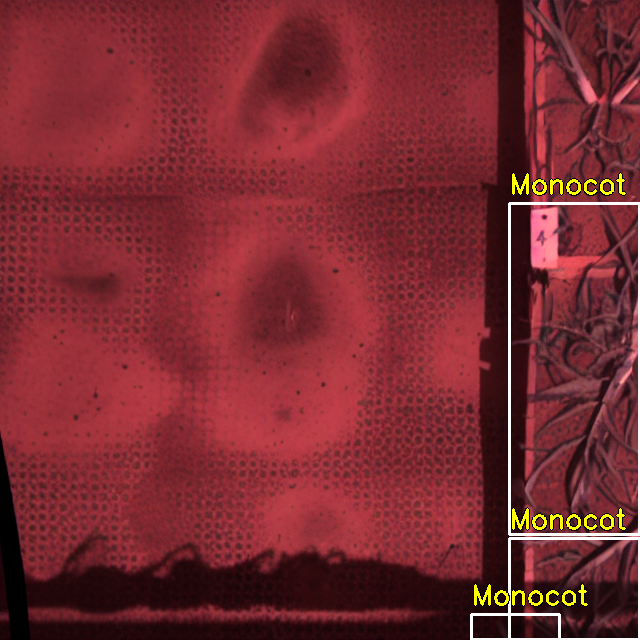}
        \subcaption*{(g)}
    \end{subfigure}
    \hfill
    \begin{subfigure}[b]{0.23\textwidth}
        \centering
        \includegraphics[width=\textwidth]{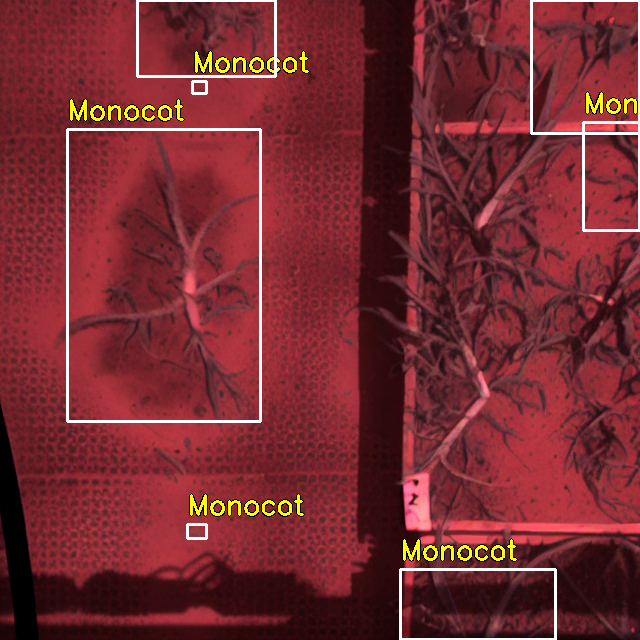}
        \subcaption*{(h)}
    \end{subfigure}
    
    \caption{Comparison of Human-Annotated and Model-Guided Annotated Images. Images (a – d) are annotated by humans, while images (e - h) are annotated using YOLOv8X }
    \label{fig:humvsml}
\end{figure}

\paragraph{\textbf{Current Limitations}} One of the primary challenges in labeling synthetic images is the lack of expert verification for model-generated annotations, which can lead to significant issues in object detection models. As shown in Figure~\ref{fig:humvsml}, model-generated annotations (e-h) exhibit several limitations compared to expert-labeled images (a-d). A major concern is inaccurate detection or misclassification. For instance, in the image (e), a dicot is incorrectly labeled as a sugar beet. Similarly, in images (e) and (f), the model incorrectly labels specific small regions, such as misidentifying species such as dicots. Additionally, it is apparent that images (f), (g), and (h) lack proper annotations, featuring either redundant labels or omissions, in contrast to the thorough human annotations in images (a-d). Specifically, in image (f), a dicot is entirely overlooked; in image (g), a monocot is missing an annotation in the upper right corner; and in image (h), certain monocots in the middle right section remain unannotated. These discrepancies may lead to confusion for object detector models during training. Therefore, relying predominantly on model-generated annotations for synthetic images, without the validation of experts, can adversely influence the efficacy of the downstream models, as evidenced by these annotation issues.

To improve the precision of annotations and provide more reliable assessments, future research should adopt an \textit{inter-(active)} annotation process, similar to the approach described by~\cite{Boysen2022}. This involves engaging domain experts to actively assess and correct model-predicted annotations, using iterative feedback to refine the results~\cite{Yao2012InteractiveOD}. Incorporating this process into our pipeline improves annotation quality and helps identify and mitigate systematic biases, ultimately leading to more accurate model performance by providing good quality annotated training data to the object detector models.

Initially, we trained the Stable Diffusion Model using the entire available real-world dataset to develop an omniscient Foundation Model for weed generation adaptable to various conditions. Subsequently, the same dataset was divided into training, validation, and test sets for the YOLO models, posing a possible risk of data leakage. Stable Diffusion Model might learn similar patterns from the test set, unintentionally introducing them into synthetic images, thus possibly biasing YOLO model training and impacting their performance. Although diffusion models generate images from random Gaussian noise (see Sect.~\ref{subsec:Diff}), the output can still reflect patterns from the original dataset~\cite{ho2020denoising, rombach2022high}. However, we've managed image generation using diverse text prompt descriptions, random seeds, and controlled generation quality with schedulers, which may mitigate bias from test data patterns. Future work will take steps to completely separate the YOLO test set from any data used in training the Stable Diffusion Model, thereby mitigating any bias that might affect the performance of the downstream model. Moreover, we plan to incrementally reduce the training data for the Stable Diffusion Model by 10\% to assess how much real data is required for training while achieving performance comparable to using the full dataset. This approach seeks to determine the volume of real-world data required to effectively train the Stable Diffusion Model and achieve comparable outcomes to train by using the entire dataset. This can potentially decrease both data collection and training costs.

\section{Conclusion and Outlook}
\label{sec:Conclusion_Future_Work} 

In the context of advancing sustainable crop protection, leveraging state-of-the-art deep learning algorithms for weed detection in intelligent weed control systems is crucial. However, a critical bottleneck hindering progress is the scarcity of high-quality datasets with sufficient size and diversity. Obtaining these datasets is not only time-consuming but also financially demanding, presenting a major challenge to the effective deployment of AI-based technologies. While dataset augmentation is a classic technique to enhance the training outcome of data-hungry DL approaches, traditional methods often fail to achieve important properties such as realism, diversity, and fidelity all at once. 
This is however required for accurately reflecting the conditions of real-world scenarios within the synthetically created image data and, thus, for obtaining models with increased robustness. In this paper, we introduced a pipeline architecture for synthetic image generation, which integrates state-of-the-art generative and foundation AI models. 
It showcases a robust methodology for enhancing image datasets to increase the data efficiency of model training tailored to computer vision tasks, with a specific focus on weed detection. 
Leveraging cutting-edge DL-based techniques, we successfully overcome the constraints of restricted realism and diversity encountered in traditional data augmentation methods. Furthermore, the synthetic images produced through our pipeline capture the natural diversity and idiosyncrasies from the considered real-world scenario of weed detection -- a critical aspect for effectively training robust models for downstream tasks. This emphasis on realism ensures that future training datasets augmented by means of such an approach have the potential to better prepare data-driven AI-based solutions to capture the innate complexities of heterogeneous environments, ultimately enhancing their performance and adaptability.

Accordingly, our proposed methodology is not only deemed promising for increasing the amount and variability of image datasets for training machine learning models. Moreover, our observations indicate that integrating a modest proportion of synthetic data during the training of different lightweight YOLO models can substitute real-world images and enhance the performance of the considered downstream task.  On the one hand, it significantly enhances data efficiency by reducing the reliance on real-world datasets of larger size involving higher cost and labor for data collection and annotations. Moreover, this indicates the potential of our approach especially for cases where deep learning-based AI models are to be deployed on edge devices.

We see a promising use case in the integration of our method into intelligent systems architectures, e.g., the multi-level observer/controller (MLOC) architecture from the Organic Computing (OC) domain~\cite{tomforde2011observation} in general, and intelligent agricultural technology systems, such as robots or smart implements (cf. e.g.,~\cite{BOYSEN2023100363}), in particular.
Such system architectures usually include higher-level reflection layers, that monitor the performance of the underlying adaptation layers, which are in turn actively controlling a system under observation and control (SuOC). These reflection layers' task is to intervene in case of automatically detected performance drops of the system by means of triggering a reconfiguration of the adaptation layers below (in turn indirectly affecting the adaptation policy of the SuOC), e.g., through switching inference models or updating their active knowledge bases~\cite{stein2021reflective}.
Integrating a reliable training data generation pipeline into a reflection layer, would allow self-learning adaptive systems to specifically counter identified knowledge gaps~\cite{Stein2018} through on-demand online training based on synthetic data. 
By using our method, the detection of knowledge gaps can be implemented in a reactive (e.g., triggered by low confidence estimates for detected weeds in a particular scene) as well as a proactive (e.g., triggered through continually creating synthetic scenes and in parallel check the active model's predictive accuracy) manner. 
In combination with continual learning techniques, these systems can at the same time prevent catastrophic forgetting of previously learned knowledge.
For the related agricultural use case of detecting and segmenting field vegetables in an automated field monitoring system, an initial version of such a self-reflective system architecture has previously been proposed by Lüling et al.~\cite{luling2022context}. In future work, we intend to dive deeper into the remaining aspects as outlined above.

\subsubsection*{Acknowledgements.}
\label{subsec:ack}
This research was conducted within the scope of the project ``Hochleistungssensorik für smarte Pflanzenschutzbehandlung (HoPla)" (FKZ 13N16327), and is supported by the Federal Ministry of Education and Research (BMBF) and VDI Technology Center on the basis of a decision by the German Bundestag.
\bibliographystyle{elsarticle-num-names} 

\bibliography{ref.bib} %

\end{document}